# Hydrogen production from blended waste biomass: pyrolysis, thermodynamic-kinetic analysis and AI-based modelling

Sana Kordoghli[1,2] Abdelhakim Settar[3], Oumayma Belaati[1], and Mohammad Alkhatib[4], Khaled Chetehouna[3], Zakaria Mansouri[5]

[1] National School of Science and Advanced Technologies, ENSTA-Borj-Cedria, University of Carthage - Tunis, Tunisia ,

[2] Research laboratory of Environmental Sciences and Technologies, ISSTE Borj Cedria, University of Carthage - Tunis, Tunisia ,

[3] INSA Centre Val de Loire, University of Orléans, PRISME UR 4229, F-18020 Bourges, France,

[4] Clermont Auvergne INP, SIGMA Clermont, Clermont-Ferrand, France,

[5] Department of Engineering, Nottingham Trent University, Nottingham, United Kingdom

*Corresponding author: Zakaria Mansouri (zak.mansouri@ntu.ac.uk)

## Abstract

This work contributes to advancing sustainable energy and waste management strategies by investigating the thermochemical conversion of food-based biomass through pyrolysis, highlighting the role of artificial intelligence (AI) in enhancing process modelling accuracy and optimization efficiency. The main objective is to explore the potential of underutilized biomass resources like spent coffee grounds (SCG) and DS (date seeds) for sustainable hydrogen production. Specifically, it aims to optimize the pyrolysis process while evaluating the performance of these resources both individually and as blends. Proximate, ultimate, fibre, TGA/DTG, kinetic, thermodynamic, and Py-Micro GC analyses were conducted for pure DS, SCG, and blends (75% DS – 25% SCG,50%DS-50%SCG,25%DS–75%SCG). Blend 3 offered superior hydrogen yield potential but had the highest activation energy (Ea: 313.24 kJ/mol), while Blend 1 exhibited the best activation energy value (Ea: 161.75 kJ/mol). The kinetic modelling based on isoconversional methods (KAS, FWO, Friedman) identified KAS as the most accurate. These approaches work together to provide a detailed understanding of the pyrolysis process with a particular emphasis on the integration of artificial intelligence (AI). An LSTM model trained with lignocellulosic data predicted TGA curves with exceptional accuracy ($R^2$ : 0.9996–0.9998).

## 1. Introduction

The world's energy needs are evolving rapidly as we face the urgent challenges of climate change and strive to transition away from fossil fuels. The harmful impacts of burning fossil fuels—rising global temperatures, increasingly severe natural disasters, and long-term damage to ecosystems have made it clear that we need cleaner and more sustainable energy sources [1][2]. Among the promising solutions is hydrogen, a clean energy carrier that can revolutionize key industries such as heavy manufacturing, long-distance transport, and energy storage[1][3]. For instance, hydrogen has the potential to reduce carbon emissions in steel production by up to 90% [3]. Despite this promise, most hydrogen today is still produced using fossil fuels, a method known as "Gray hydrogen," which does little to address carbon emissions. Shifting to "green hydrogen," made from renewable energy sources like wind and solar, is essential but challenging, especially given the intermittency of these renewable energy supplies. In the face of these challenges, biomass emerges as a steady and reliable alternative for hydrogen production. Agricultural residues and food wastes, such as date seeds (DS) and spent coffee grounds (SCG), are often overlooked yet possess immense potential for creating sustainable energy. Not only does utilizing these resources reduce waste, but it also aligns with global efforts to transition toward cleaner energy systems. Pyrolysis provides a sustainable waste-to-resource solution, converting these biomass wastes into valuable products like bio-oil, biochar, and syngas, thereby reducing landfill burden and contributing to circular economy models. However, to fully unlock the potential of biomass for hydrogen production, it is crucial to optimize the processes involved and address the inherent complexities of biomass composition. This is where machine learning (ML) and its subsets, such as deep learning, come into play as game-changing tools. By analyzing large datasets, ML algorithms can uncover hidden patterns, predict outcomes, and optimize parameters that would be challenging to manage using traditional methods[4]. For example, neural networks excel at modelling the intricate relationships found in thermogravimetric and kinetic data. This capability allows for precise predictions of how biomass behaves during pyrolysis and its efficiency in producing energy [5]. By reducing the need for labour-intensive experiments, predictive models driven by ML streamline research processes, making bioenergy development faster and more cost-effective. Building on this foundation, the focus of this study is to explore the potential of underutilized biomass resources like SCG and DS for sustainable hydrogen production. Specifically, it aims to optimize the pyrolysis process—a proven method for producing hydrogen while evaluating the performance of these resources both individually and as blends. To gain deeper insights, the study employs a range of analyses, including compositional studies, thermogravimetric analysis (TGA) under pyrolysis conditions, kinetic and thermodynamic evaluations, and pyrolysis tests. These approaches work together to provide a detailed understanding of the pyrolysis process and an investigation to prove which biomass sample ensures a more efficient and practical hydrogen production. To complement this experimental work, the study incorporates predictive modelling powered by Long Short-Term Memory (LSTM) neural networks. These advanced models are used to forecast mass loss curves from TGA data, offering a time- and cost-efficient alternative to traditional methods. By optimizing the pyrolysis process through these models, the study demonstrates how artificial intelligence can accelerate and enhance bioenergy research.

Machine learning techniques, particularly neural networks, have been increasingly applied in thermal analysis and pyrolysis research due to their ability to accurately model complex relationships between parameters and outcomes. Studies using traditional approaches like Random Forest (RF) and Support Vector Regression (SVR) have demonstrated their effectiveness in predicting reaction kinetics and optimizing operational parameters [6],[7],[8],[9]. However, artificial neural networks (ANNs) have gained prominence for achieving superior performance in predicting pyrolysis outcomes, such as product yields and reaction kinetics[10],[11]. For example, Balsora et al.[12] used ANNs to predict product yields with $R^2$ values around 0.97, and Kartal and Özveren[13] applied ANNs to estimate kinetic parameters with $R^2$ values over 0.96. While most studies rely on multi-layer perceptrons (MLPs), which work well for static data, the sequential nature of TGA data makes Recurrent Neural Networks (RNNs) and Long Short-Term Memory (LSTM) networks more suitable. LSTMs effectively capture temporal dependencies and address challenges like the vanishing gradient problem, making them ideal for modelling weight-loss patterns during pyrolysis[14],[15].

The existing literature demonstrates considerable advances in the application of AI and machine learning (ML) techniques for thermogravimetric analysis (TGA) modelling of biomass pyrolysis. However, to date, no published studies have specifically applied these tools to coffee grounds, date seeds, or their blends. This notable gap underscores a timely opportunity for targeted, high-impact research leveraging recent methodological breakthroughs in AI/ML to explore the pyrolytic behaviour of these underutilized biomass resources. In the same underlying ideas, this study combines experimental analysis, innovative blending strategies, and advanced modelling to present a comprehensive approach to biomass-based bioenergy production. By improving pyrolysis efficiency and leveraging underutilized resources, it supports the global shift to sustainable energy systems and paves the way for a cleaner and more circular hydrogen economy. Therefore, the objectives of this article consist of assessing the potential of these biomass wastes for sustainable energy and chemical production; developing artificial intelligence models to predict TGA output curves, and finally comparing experimental results with model predictions and existing literature. The comprehensive methodology used in this work reveals promising outcomes, particularly in the application of machine learning model where high predictive accuracy were achieved even for blended food waste biomasses.

## 2. Materials and Methods

### 2.1. Samples preparation

The biomass materials utilized in this study were Spent Coffee Grounds (SCG) and Date Seeds (DS) ( Figure 1) . The SCG was sourced from a Tunisian coffee shop, while the DS was collected from an industry specializing in processing Phoenix dactylifera dates. Before conducting any experiments, both biomass samples underwent a drying process to remove residual moisture. The SCG was dried using a Memmert UF55 drying oven set at 40°C for a duration of 16 hours. Similarly, the DS was dried in an oven at 105°C, then cooled and ground using a Retsch MM400 ball mill. After grinding, the DS samples were stored in moisture proof containers to ensure their preservation for future use. Once prepared, both types of biomasses were stored in airtight containers to maintain their condition until further analysis.

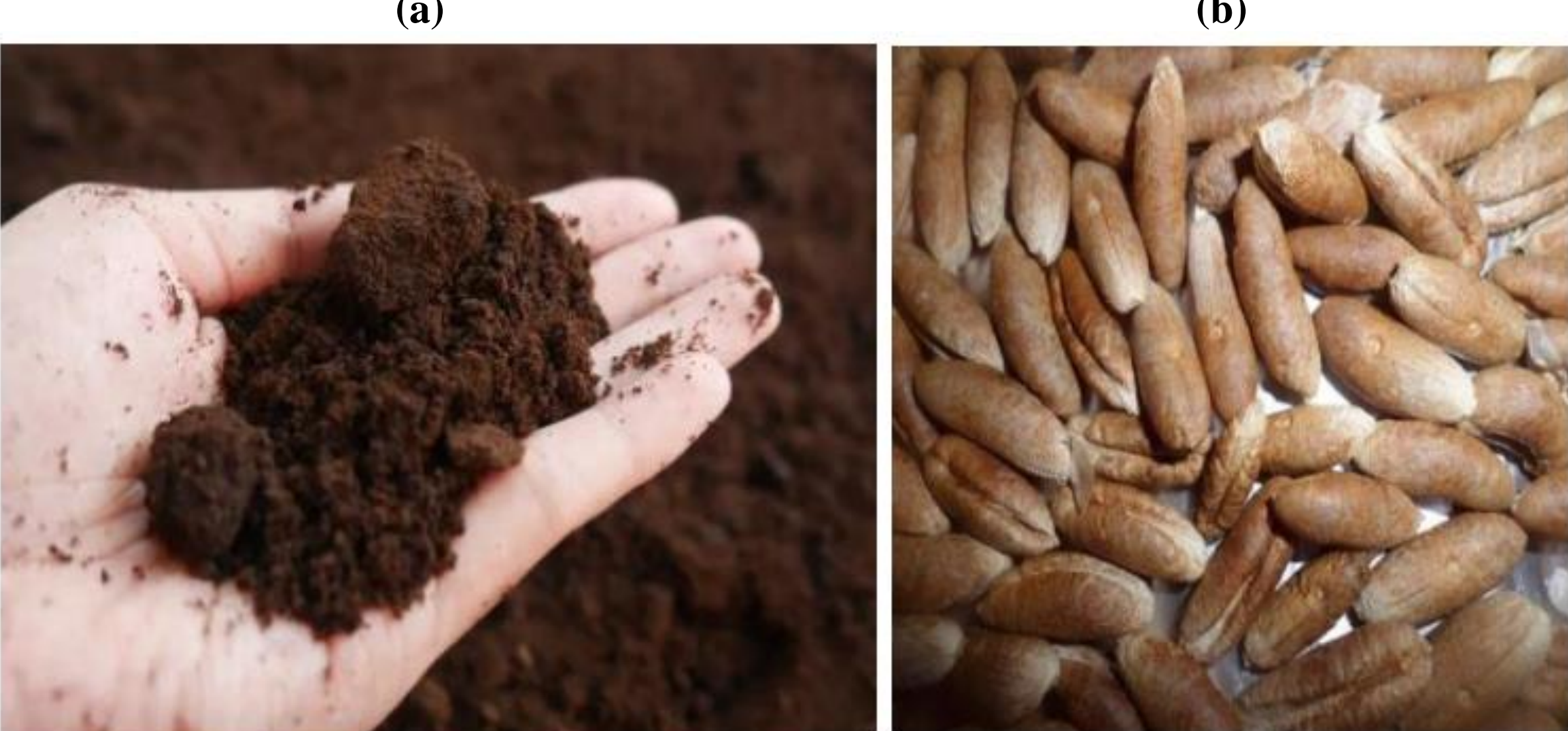

**Figure 1.** Investigated biomass feedstocks: (a) SCG and (b) DS.

### 2.2. Experimental facilities

The pyrolysis experiments were conducted in a laboratory-scale fixed-bed reactor at final temperatures of 650°C, with a heating rate of 10°C/min and under atmospheric pressure. The fixed bed is composed of an inert solid material, covered with a small layer of glass wool to prevent temperature loss and guarantee thus a good conductivity. The biomass sample was placed in the reactor vessel. A Tedlar bag was attached to the reactor outlet to collect the gas. The reactor's outlet valve (for gas release) and the Tedlar bag were opened, while an inert atmosphere was ensured by injecting nitrogen at a specified flow rate (50 ml/min for 15 minutes). The heating rate was set to 10°C/min. For the initial tests, a temperature of 650°C was selected, with monitoring performed using PicoLog software.

Ultimate analysis is performed to identify the elemental composition of the samples. It was performed employing a CHNOS elemental analyser (Flash EA 1112 Series) that supplies percentage compositions for carbon, hydrogen, nitrogen, and sulphur, with oxygen being obtained by subtraction. The thermogravimetric study was performed from ambient temperature to 850°C employing a SETARAM ThermysOne-TG-DSC, which is capable of reaching temperatures up to 1600°C, under an inert atmosphere provided by a nitrogen gas flow rate of 50 ml/min. Finally, Micro gas chromatography (Micro-GC), a miniaturized version of gas-liquid chromatography (GLC), is a highly efficient and precise technique for separating and analyzing volatile components in mixtures. When combined with pyrolysis (Py-GC), micro-GC becomes an invaluable tool for analyzing the volatile products generated by the thermal degradation of complex organic materials.

### 2.3. LSTM-based prediction of TGA data

After discussion results on the physicochemical, thermal degradation and kinetic behaviours, the data outcomes are employed to feed the LSTM-based approach. To remind, it aims to predict TGA curves for biomass blends, leveraging their ability to process sequential data and capture complex thermal degradation trends. The model was developed and tested using Python 3, utilizing libraries such as TensorFlow and Keras for building the LSTM model, Scikit-learn for pre-processing, and Keras Tuner for hyper parameter optimization. Tools like Matplotlib were employed for result visualization, while RandomSearch was used to identify the optimal configuration for the LSTM architecture. By splitting the data into training, validation, and test subsets, the model's performance was rigorously evaluated on unseen data, ensuring robust and generalizable predictions

### 2.3.1. Data pre-processing and feature engineering

The data obtained from TGA experiments on both pure and blended biomass samples are used to feed the deep learning procedure. Each sample was analysed across a range of heating rates up to a final temperature of 900°C. This extensive data provided insights into the thermal degradation patterns of both individual and blended samples. The specific datasets included:

- Spent Coffee Grounds (SCG): TGA data collected at heating rates of 5, 10, 15, and 20°C/min;
- Date Seeds (DS): TGA data collected at heating rates of 5, 10, 15, and 20°C/min;
- Blend 1 (75% DS, 25% SCG): Data collected at heating rates of 5, 10, 15, and 20°C/min;
- Blend 2 (50% DS, 50% SCG): Data collected at heating rates of 5, 10, and 15°C/min;
- Blend 3 (25% DS, 75% SCG): Data collected at heating rates of 5, 10, and 15°C/min.

These various datasets allowed us to observe and model thermal degradation trends across different compositions and heating rates, which was essential for building an accurate predictive model. The raw TGA data collected was complete and did not contain any missing values, as the TGA instrument generated a full dataset for each measurement. Furthermore, raw data were used without applying noise reduction or smoothing techniques to preserve reliability. This allowed the model to learn directly from the inherent variations within the data, which is expected to support generalization in prediction tasks. Concerning the model training, two dataset versions were created, each designed to capture different layers of detail:

- Model 1: Primary Dataset: This simpler dataset included core features—DS %, SCG %, heating rate (°C/min), sample temperature (°C), and mass % (the target variable representing the sample's remaining mass percentage during heating). These features were selected to provide a straightforward representation of the sample composition, heating conditions, and the resulting mass loss over time.
- Model 2: Extended Dataset with Lignocellulosic Composition: To deepen the model's understanding of biomass thermal behaviour, this dataset included additional features representing the three main lignocellulosic components—Cellulose %, Hemicellulose %, and Lignin %.

For Model 2, decomposition characteristics of lignocellulosic components over different temperature ranges were considered, each affecting mass loss differently as the temperature increases:

- Cellulose % decomposes rapidly between 315 and 405°C, causing significant mass loss in this range.
- Hemicellulose % begins decomposing at lower temperatures (around 225–325°C), contributing to initial mass loss.
- Lignin % decomposes slowly over a wider temperature range (160–900°C), resulting in gradual mass loss that extends throughout the TGA process.

Thus, to accurately reflect how each of these lignocellulosic components breaks down at different temperatures, the proportions of Cellulose %, Hemicellulose %, and Lignin % were adjusted dynamically with temperature changes during the TGA tests. By incorporating these temperature-dependent adjustments in lignocellulosic composition, this extended dataset captured the dynamic nature of each biomass component's decomposition. This allowed the model to learn more about the complex relationship between temperature, heating rate, composition, and mass loss, enhancing its ability to accurately predict TGA curves.

Given the sequential nature of TGA data, with mass loss occurring over a temperature range, LSTMs offer a robust solution by managing long-term dependencies and efficiently processing complex temporal patterns. By addressing gradient issues and incorporating memory cells and gates, LSTMs provide the capacity to capture the intricate relationship between sample composition, heating rate, and mass loss across varying temperatures, making them an ideal choice for predictive modelling in this study.

### 2.3.1. Model training, and hyperparameter tuning

The dataset, consisting of 14,875 data points was divided into three subsets. Seventy percent (70%) of the data was used for training the model ensuring the model's ability to generalize to new unseen data, while 15% was reserved for validation. This subset plays a critical role in monitoring the model's performance during training. The remaining 15% of data was dedicated to testing. This leads to evaluate the model's ability to generalize by providing feedback on how well it performs on data that was not used in the training process. Furthermore, the validation loss is tracked during training, and if the model starts overfitting, the training is stopped early, helping to prevent overfitting. Once the training and validation phases were complete, the model was evaluated using a completely unseen dataset, which was not part of either the training or validation sets. This test dataset consisted of data from Blend 1, evaluated under different heating rates: 15°C/min, 10°C/min, and a completely new heating rate of 25°C/min. This final test phase assesses the model's generalization capability and its performance on truly new data. The entire model training, validation, and testing process was conducted using Google Colab, which provides open access to computational resources such as GPUs. with rapid execution of deep learning tests.

To optimize the model, Keras Tuner was used for hyperparameter tuning. Hyperparameters such as the number of LSTM layers, number of units in each layer, dropout rates, activation functions, and the optimizer were explored systematically. Furthermore, in order to forecast the subsequent value in the time series, the applied LSTM model uses the input sequences of 20 prior TGA measurements. This look-back window was part of the hyperparameter tuning process and was considered as a hyperparameter. In addition, the RandomSearch method was employed, which searches through the hyperparameter space by randomly selecting combinations and evaluating their performance (Table 1 and 2).

**Table 1.** Hyperparameters Configuration.

| Hyperparameters | Range Tested |
|---|---|
| Learning Rate | [0.0001,0.01] |
| Batch Size | [32,64] |
| Epochs | [10,50] with step=10 |
| Dropout rate | [0.1,0.5] with step= 0.1 |
| Hidden Units | [64, 256] with step=32 |

This approach was chosen for its efficiency in finding well-performing configurations, especially with a large number of hyperparameters to explore. The hyperparameter tuning process involved training multiple models, each with a different combination of hyperparameters, as is mentioned in Tables 1 and 2. Finally, the model that achieved the lowest validation loss was selected for further evaluation (Figure 2).

**Table 2.** Layers and Activation Functions.

| Layer | Number of layers Tested | Activation Functions /Optimizers Tested |
|---|---|---|
| Input Layer | 4 and 7 | None |
| LSTM Layers | From 1 to 3 | ReLU, Sigmoid and Tanh |
| Dense Layer | 1 | None |
| Optimizers | - | Adam, SGD and RMSprop |

The considered architecture of the neural network comprises: (1) The input layer (R4) which processes the TGA data, followed by (2) two hidden layers (R9 and R12) comprising the LSTM cells which lead to learn an retain temporal sequences of biomass pyrolysis behaviour. The present architecture allows to capture both short-term and long-term dependencies during the thermal decomposition patterns. Finally, (3) an output layer (R1) to generate the predicted mass loss evolution. This layer serves to compare predictive data against experimental ones, thereby evaluating model's prediction accuracy. Figure 2 depicts the optimal hyperparameters for each model.

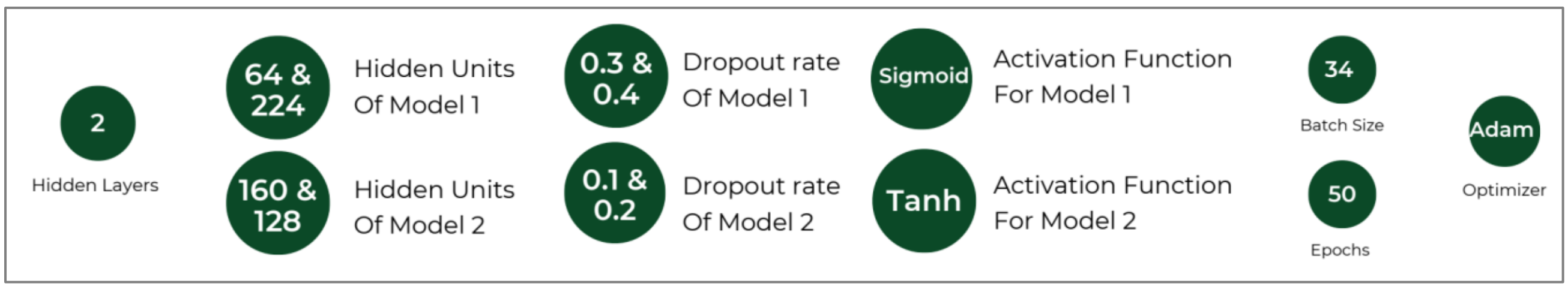

**Figure 2.** Optimal hyperparameters for each model.

To evaluate the performance of the model, several error metrics were used: Mean Absolute Error (MAE) to quantify the average of the absolute differences between actual values and their corresponding predicted values; Root Mean Squared Error (RMSE) offering an indication of how closely the model's predictions align with observed data; and finally, the R-Squared ($R^2$) coefficient.

## 3. Results and Discussion

In this section, the results are organised as follows: First, extensive physicochemical characterisation was performed to explore the feedstocks properties. Subsequently, the thermal degradation behaviour was investigated through TGA/DTG to exhibit the mass loss under pyrolysis conditions. Kinetic modelling using isoconversional methods was then applied to determine activation energies and reactional mechanisms. Finally, basing of the aforementioned findings, a predictive approach using LSTM model was performed to optimise pyrolysis process parameters. The results on the pure samples (SCG and DS) are first discussed in each subsection, then the investigated three blends: Blend 1 (75% DS & 25% SCG); Blend 2 (50% DS & 50% SCG) and Blend 3 (25% DS & 75% SCG).

### 3.1. Physicochemical characterization: Proximate, ultimate, thermal & fibre analysis

The proximate, ultimate, and thermal analysis of Date Seeds (DS) and Spent Coffee Grounds (SCG) (Table 3) shows values that align closely with those reported in the literature [16], [17],[18],[19],[20],[21],[22]. Both biomasses exhibit characteristics favourable for bioenergy applications, including suitable volatile matter (VM) and fixed carbon (FC) content for reactivity and char production. Their ultimate analysis highlights comparable carbon, hydrogen, and oxygen contents, with SCG showing slightly better energy efficiency due to lower oxygen content. Additionally, their high heating values confirm strong potential for producing energy-rich gases like hydrogen during pyrolysis [23]. These results validate DS and SCG as promising feedstocks for biofuel production, comparable to other agricultural residues [24] ,[25].

**Table 3.** Proximate, ultimate, and thermal analysis of date seeds and spent coffee grounds

| **Samples** | **Proximate Analysis** | | | **Ultimate Analysis** | | | | **HHV (MJ/Kg)** |
|---|---|---|---|---|---|---|---|---|
| | Ash (%) | VM (%) | FC (%) | C (%) | H (%) | N (%) | O (%) | |
| DS | 1.2 | 77.6 | 21.2 | 47.9 | 6.6 | 0.9 | 44.5 | 19.9 |
| SCG | 1.8 | 77.9 | 20.3 | 50.4 | 6.9 | 2.4 | 40.3 | 21.4 |

The lignocellulosic content of the two pure samples used in this study, along with the fibre analysis results from other studies on the same samples, is presented in Table 4 to facilitate comparison and analysis of their potential for bioenergy, particularly hydrogen production.

It is evident that the lignocellulosic content of DS and SCG falls within the same range as the results reported in similar studies [19],[23],[26],[27] . The biomass samples of date seeds (DS) and spent coffee grounds (SCG) both show promising potential for biofuel production based on their lignocellulosic composition. The high hemicellulose content in DS is particularly favourable, as hemicellulose contributes to a high biofuel potential, as highlighted by Sorek et al. [28]. Although DS's cellulose content is moderate, it still supports biofuel production. The lignin content of 25.7% is not excessively high, meaning it won't significantly hinder the breakdown of cellulose and hemicellulose [29], [19]. Similarly, SCG also shows favourable attributes for biofuel production. SCG's higher cellulose content enhances its potential, and although the hemicellulose content is moderate; it still plays a significant role in contributing to biofuel yield. Both samples' lignin content at 25% is moderate and unlikely to pose significant barriers to biofuel production.

**Table 4.** Fibre analysis of date seeds and spent coffee grounds

| **Samples** | **Fibre analysis** | | |
|---|---|---|---|
| | Cellulose (%) | Hemicellulose (%) | Lignin (%) |
| DS | 22.5 | 48.2 | 25.7 |
| SCG | 32 | 35 | 25 |

## 3.2. Thermogravimetric and kinetic analysis

### 3.2.1. TGA/DTG of pure Samples

The mass loss of spent coffee grounds (SCG), shown by Figure 3, occurs mainly between 250°C and 500°C. The DTG curves show two significant peaks and a smaller peak around 100°C, which is due to moisture evaporation during the drying stage. The first major peak corresponds to the decomposition of hemicellulose (35%), and the second to cellulose (32%). This confirms that hemicellulose is more reactive and decomposes earlier than cellulose. The pyrolysis of SCG can be divided into three stages. The first stage (up to 100°C) involves moisture loss and the breakdown of some small molecules (extractives). The second stage, from 220°C to 500°C, represents the active pyrolysis zone, where hemicellulose and cellulose are. The third stage, related to the slow decomposition of lignin, is less evident in SCG, likely because lignin, in this case, decomposes over a much broader range (160°C–900°C)[30] .

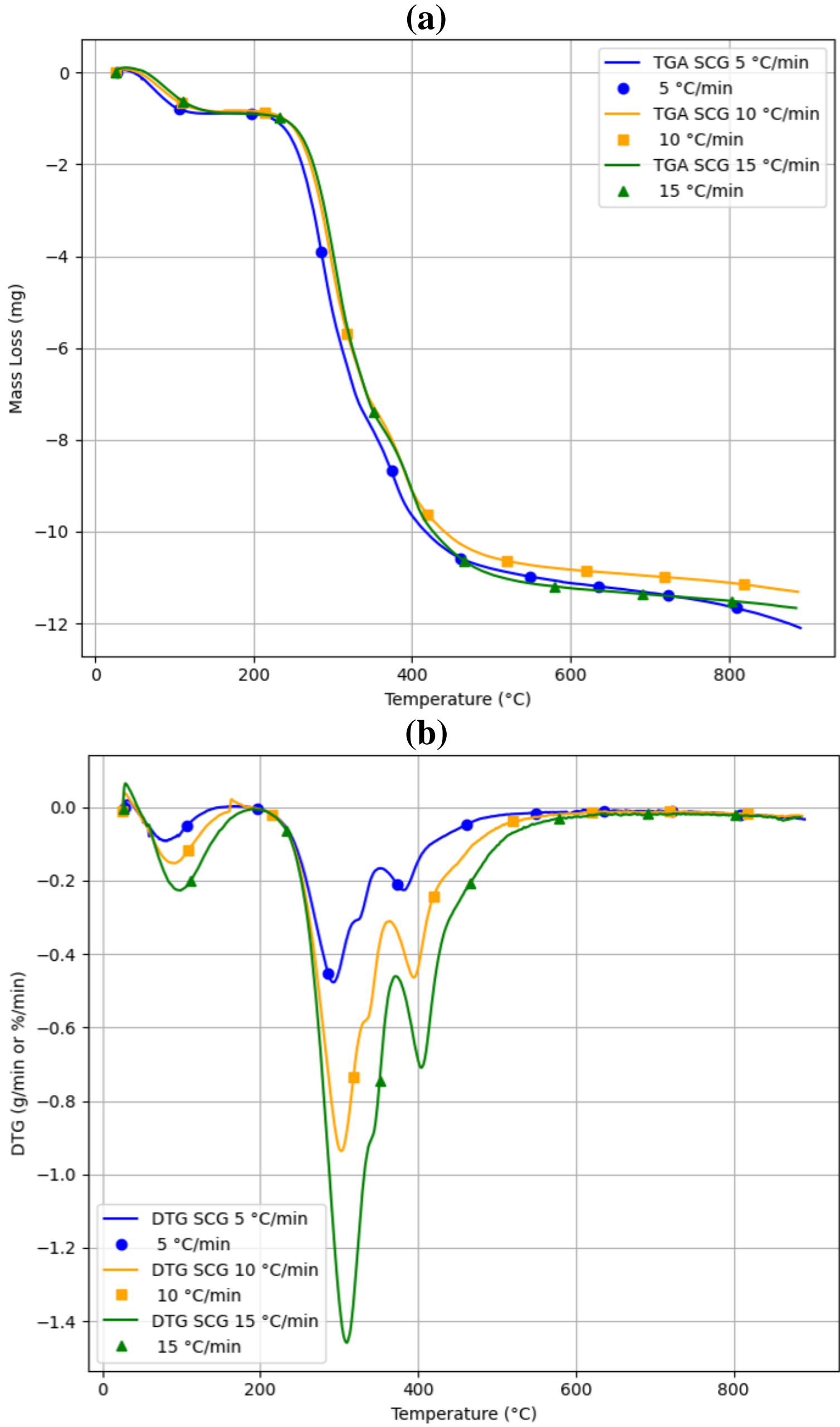

**Figure 3.** SCG mass loss as function of temperature: (a) TGA; (b) DTG.

Similarly, the thermal decomposition of date seeds (DS), shown in Figure 4, begins near 200°C, with most of the weight loss occurring by 500°C. The first major phase (200°C–350°C) is due to hemicellulose decomposition (48.2%), while the second phase (350°C–450°C) corresponds to cellulose breakdown (22.5%). Similar findings were reported by Fadhil et al.[31].

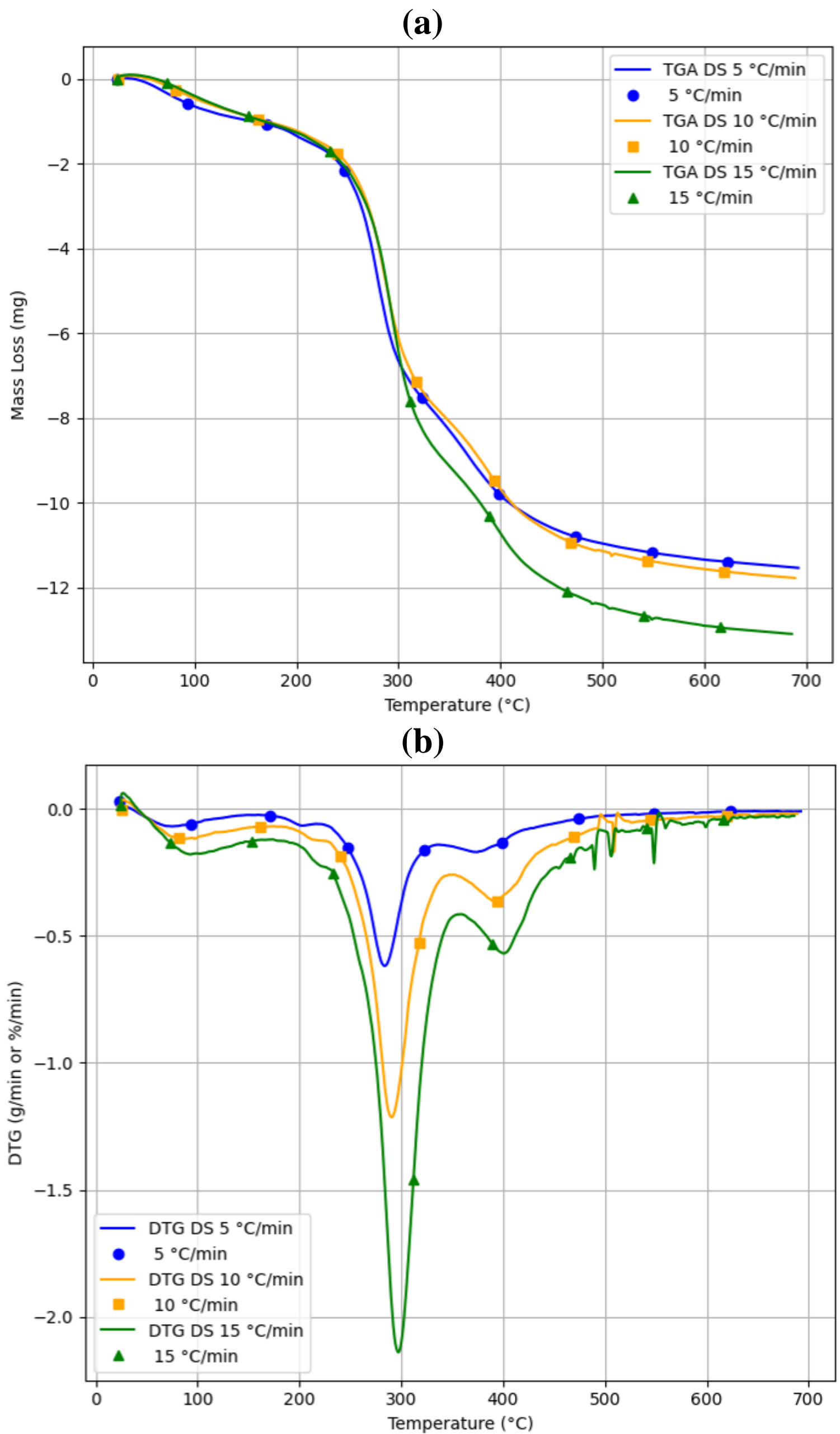

**Figure 4.** DS mass loss as function of temperature: (a) TGA; (b) DTG.

### 3.2.2. TGA/DTG of blended samples

DTG curves of Blend 1 (Figure 5) represented three different mass loss stages in each heating rate studied, as is also seen in the case of the pure samples. In general, the first stage corresponds to loss of moisture [32]. The temperature interval for this stage was below 160°C. The second stage, known as decomposition of hemicellulose, is represented by the lowest peak in every DTG curve of Figure 5; this decomposition mainly occurred in the range of 200°C - 360°C. The third stage should be the decomposition of cellulose, which occurs in the range of 360°C to 400°C for the lower heating rates of 5°C/min and 10°C/min. However, for blend 1, the decomposition of cellulose at 15°C/min isn't represented by a noticeable peak. The absence of this peak can be attributed to several factors. For instance, at higher heating rates, the decomposition of different components, such as hemicellulose, cellulose, and lignin, occurs over broader and overlapping temperature ranges due to rapid heat transfer, which can obscure distinct peaks, including that of cellulose [33]. In contrast, slower heating rates allow for better resolution between decomposition stages, providing more time for heat transfer, enabling clearer separation of decomposition events, and making the cellulose peak more visible in the DTG curves at 5°C/min and 10°C/min. The lignin decomposition isn't also observable in the DTG curves of all heating rates of this blend, which can be attributed to the broader range and the slow decomposition of this component, as it is seen in the case of the pure samples.

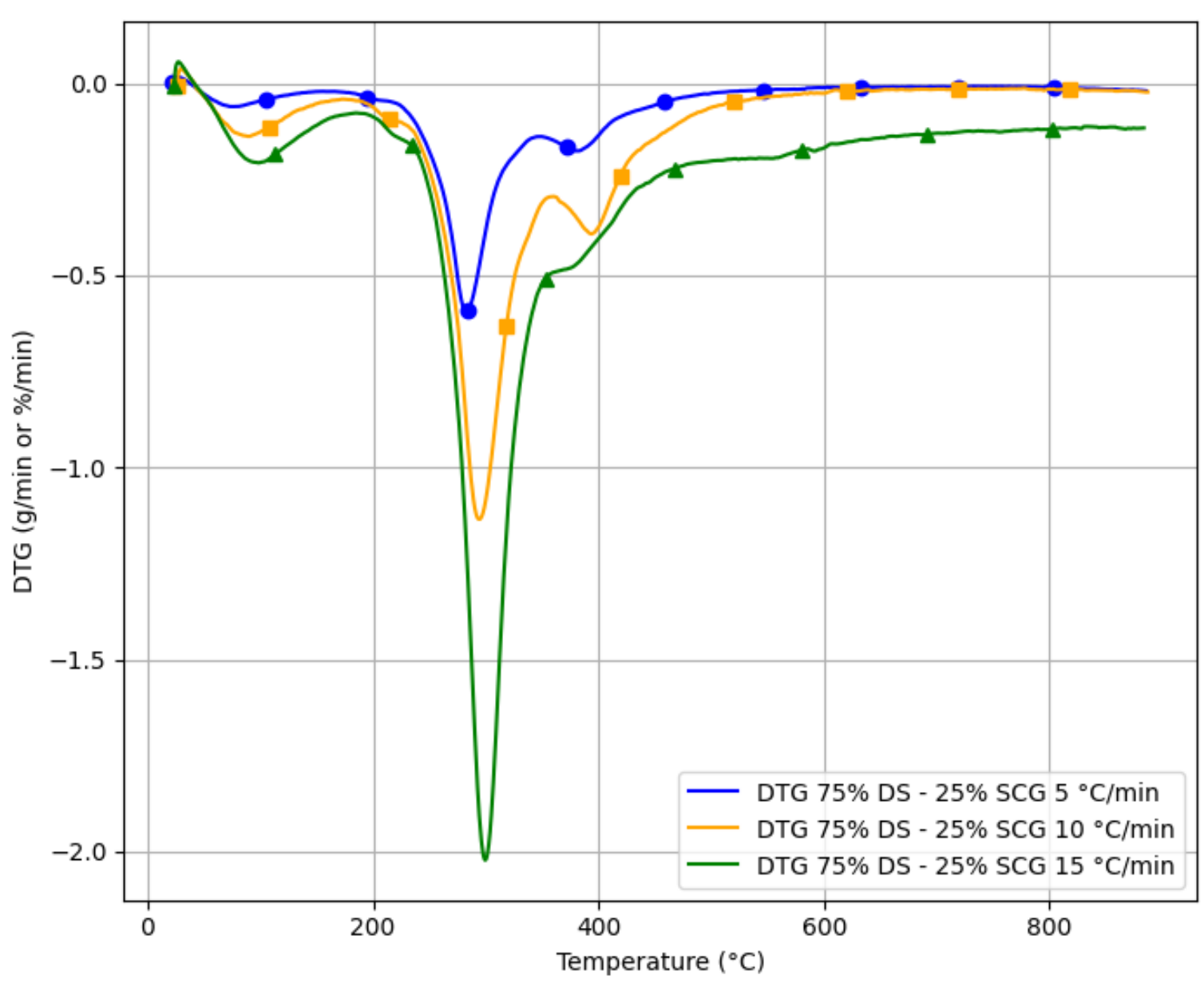

**Figure 5.** DTG evolution as function of temperature for Blend 1 (75% DS & 25% SCG).

The DTG of the second blend (50% DS and 50% SCG) (Figure 6) presents the same decomposition stages as Blend 1. The third stage, which represents the cellulose decomposition, was only observable in the DTG curve at 5°C/min and 15°C/min. However, this stage wasn't represented by a peak at the heating rate of 10°C/min. The potential explanation of this phenomenon, specific to Blend 1 and 2, is that in Blend 1 and at 15°C/min, the cellulose decomposition is potentially masked by the rapid degradation of other components (such as hemicellulose and lignin) or by complex interactions between components in the blend. At slower rates (5°C/min and 10°C/min), the separation of decomposition events becomes clearer, allowing the cellulose peak to appear more distinctly. However, in Blend 2 at 10°C/min, the cellulose peak may be masked because the decomposition of hemicellulose or lignin occurs at a similar temperature, and at this heating rate, there may not be enough separation between the different decomposition events. However, at 15°C/min, the decomposition of cellulose might occur faster and more distinctly, as there is less interference from other components, allowing the peak to be observed.

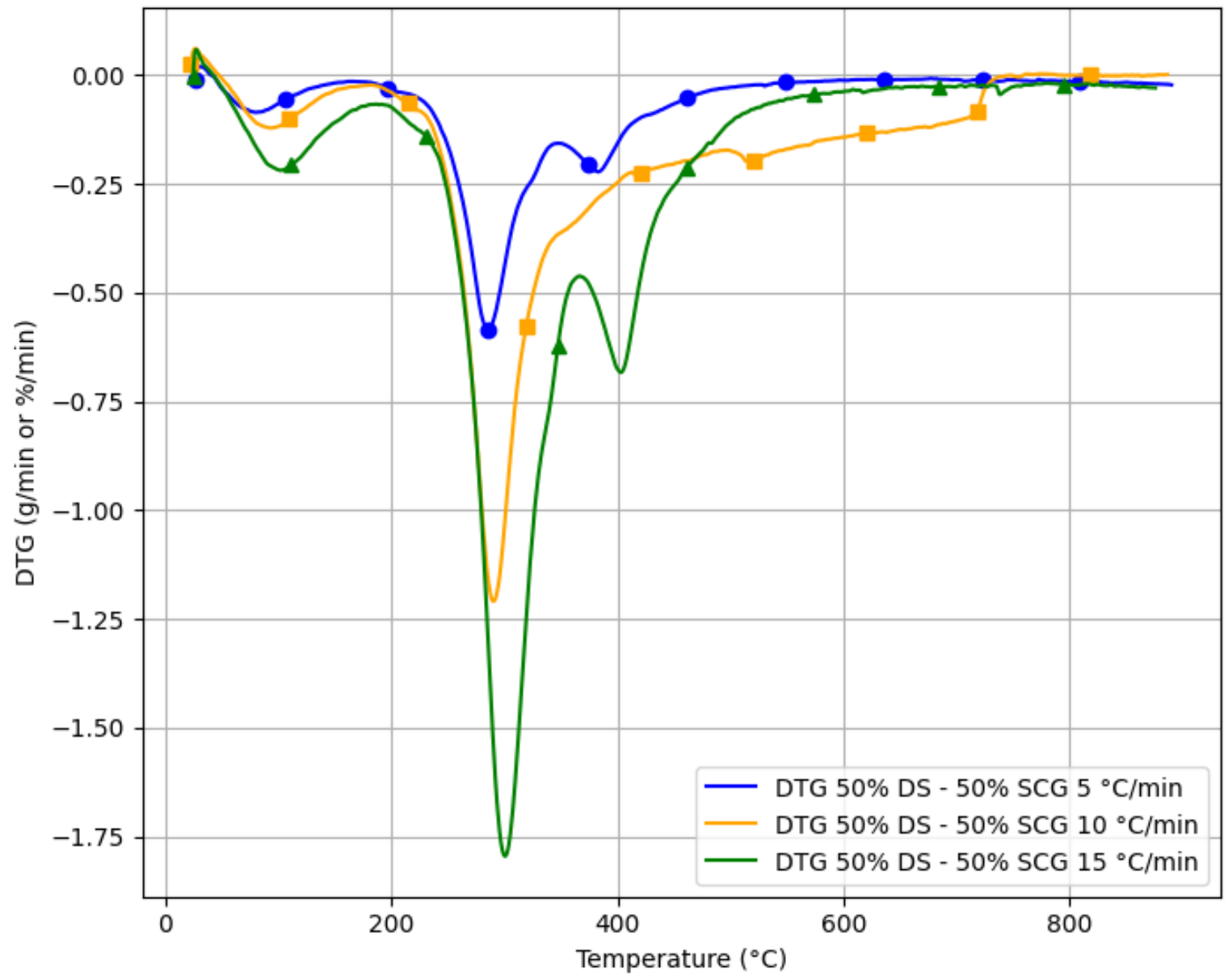

**Figure 6.** DTG evolution as function of temperature for Blend 2 (50% DS & 50% SCG).

The thermal behaviour of Blend 3, is represented by Figure 7. In this blend, all three stages of decomposition are well observed and represented by distinct peaks.

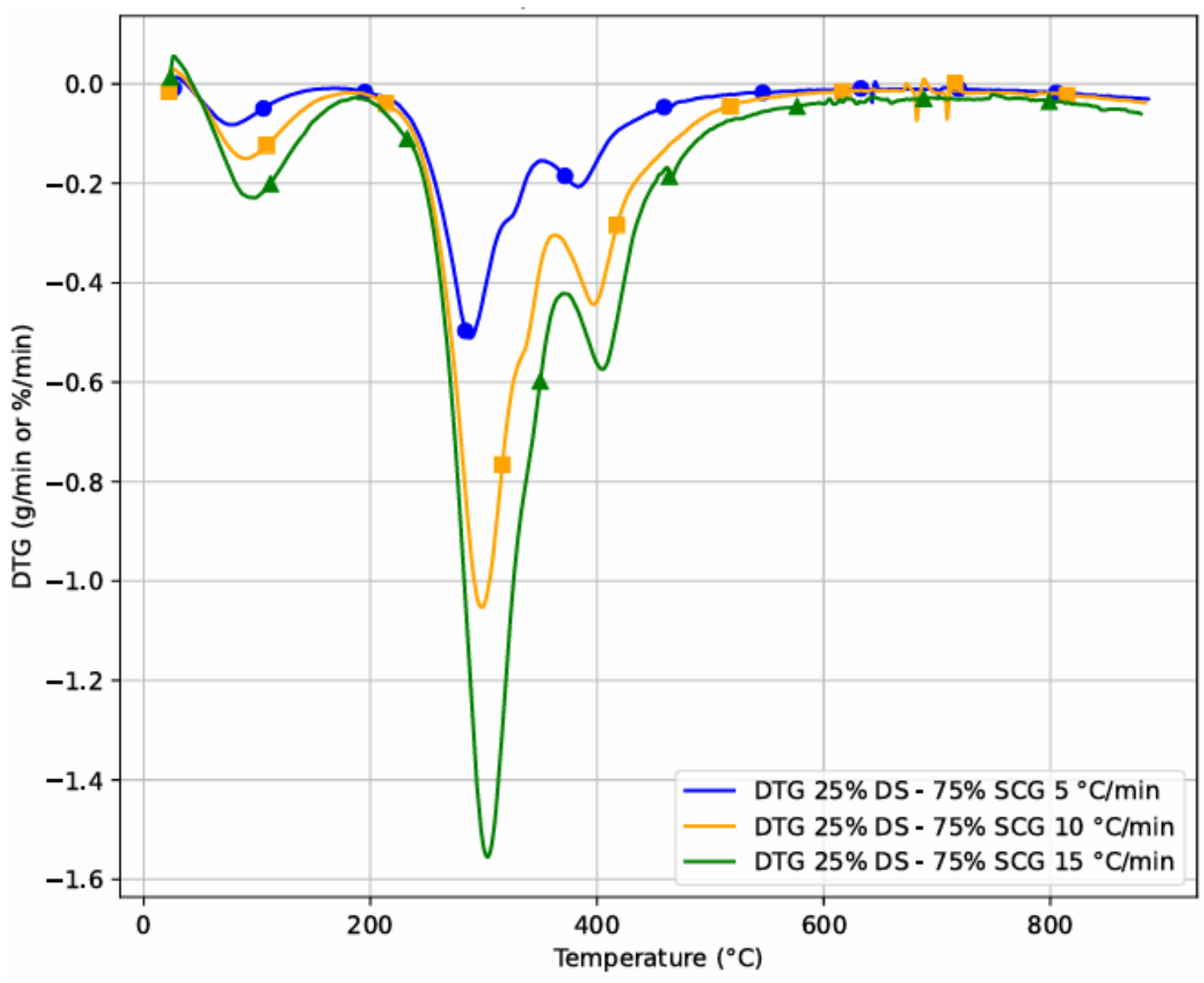

**Figure 7.** DTG evolution as function of temperature for Blend 3 (25% DS & 75% SCG).

The following figures present the DTG graphs of the three blends at a fixed heating rate (15°C/min), comparing them to the DTG graphs of the pure samples. In Figure 8, the moisture loss peak is the same for SCG and the three blends. However, the hemicellulose decomposition peak of SCG is the highest compared to all the blends, and the lower the SCG content in the blend, the lower the hemicellulose peak. Meanwhile, the cellulose decomposition peak is the lowest for pure SCG, and the lower the SCG content, the higher the cellulose decomposition peak.

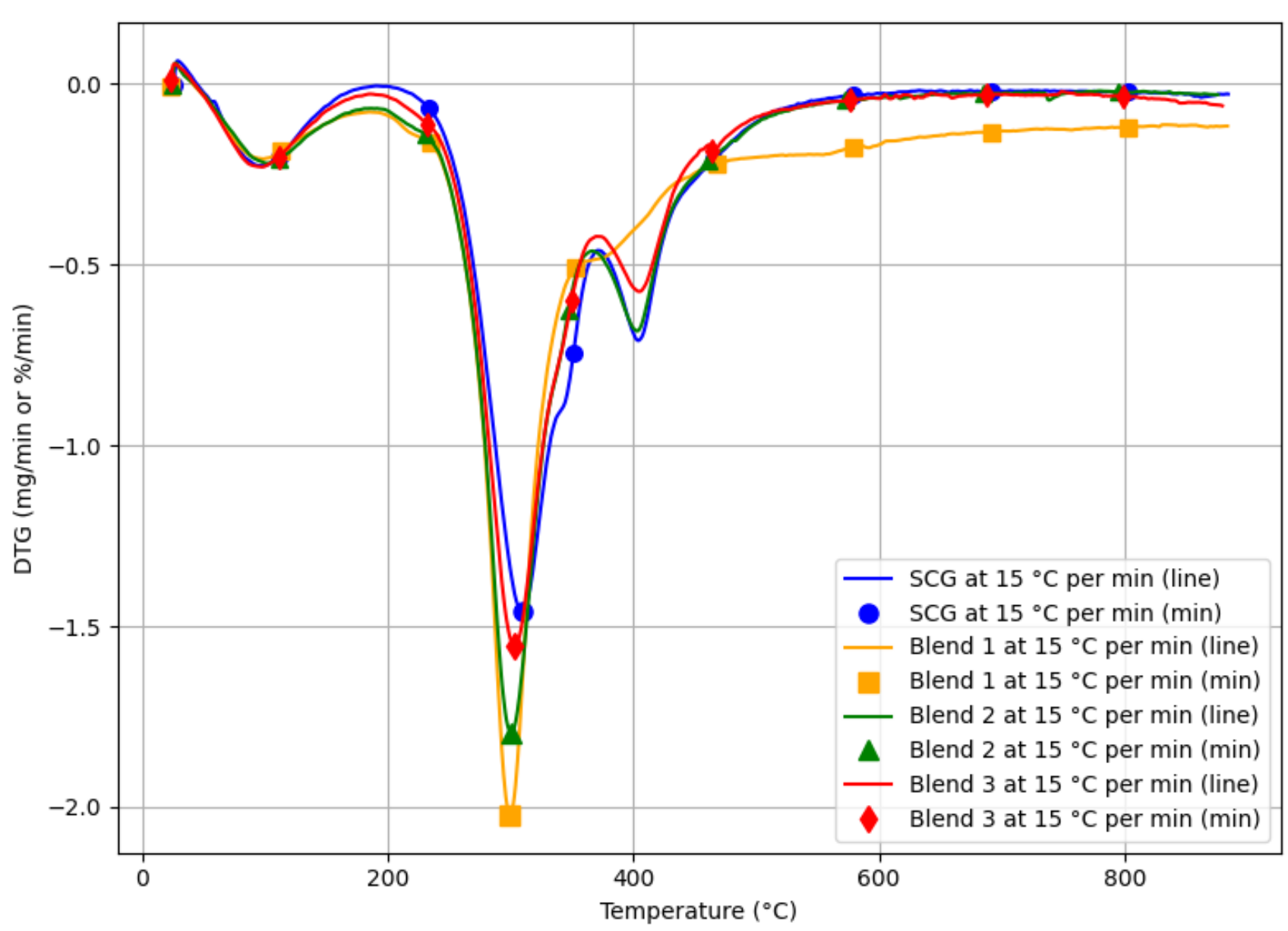

**Figure 8.** DTG evolution as function of temperature (pure SCG vs. Blends).

In Figure 9 The moisture peaks are approximately aligned. However, the hemicellulose decomposition peak of DS is the lowest among all blends, while the cellulose peak is higher in pure DS compared to the blends. This is logical since DS contains a higher hemicellulose content, meaning the higher the DS content, the higher the intensity of the hemicellulose decomposition peak.

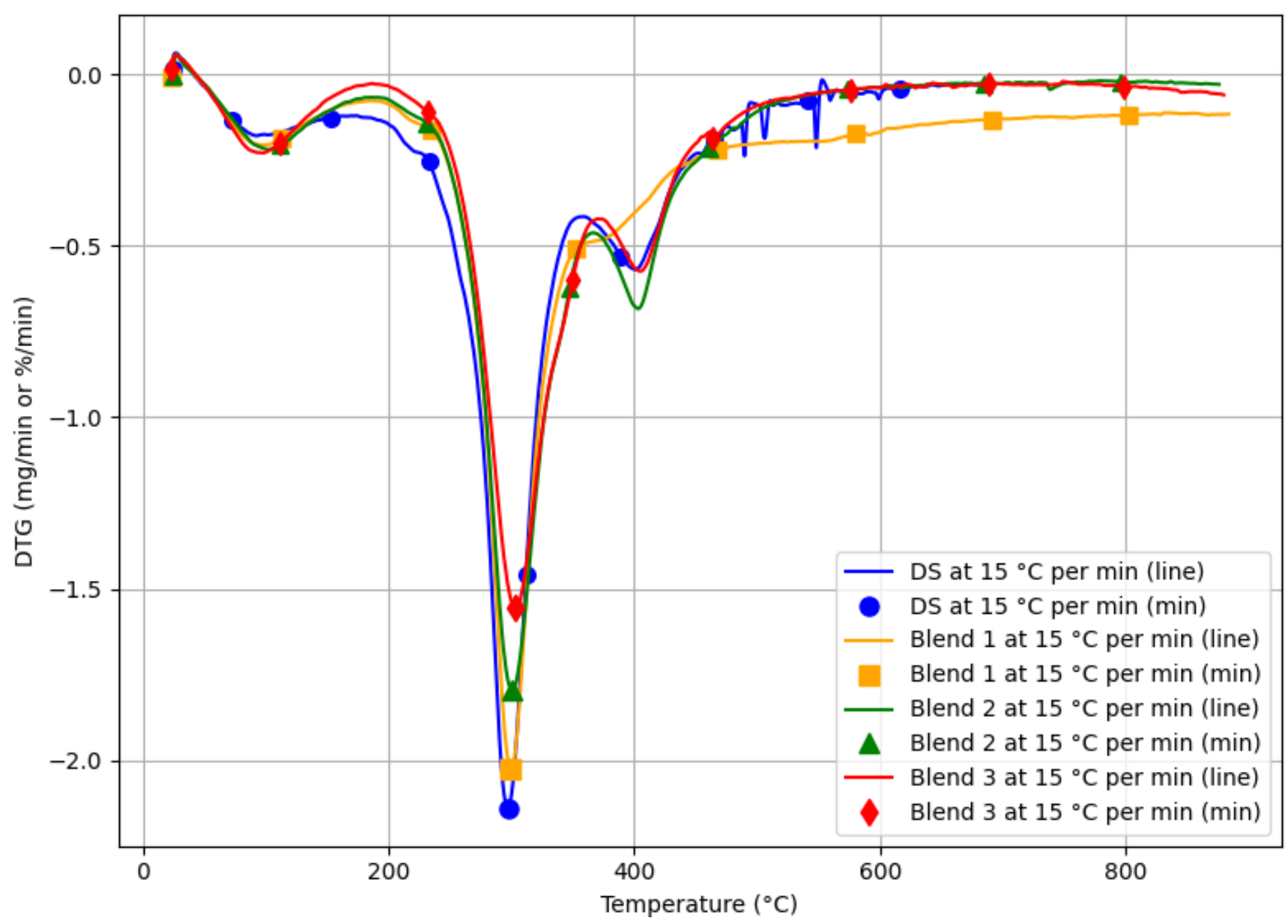

**Figure 9.** DTG evolution as function of temperature (pure DS vs. Blends).

### 3.2.3. Kinetic analysis of pure Samples

Across the three models (Friedman, KAS, and FWO) in Table 5 and Figure 10, the average Ea values are relatively close—296.03 kJ/mol (Friedman), 284.86 kJ/mol (KAS), and 279.91 kJ/mol (FWO) indicating they provide consistent results for the energy barrier of SCG pyrolysis.

**Table 5.** Comparison of Kinetic Parameters from Friedman, KAS, and FWO Models for pure SCG.

| α | Friedman Model | | | KAS Model | | | FWO Model | | |
|---|---|---|---|---|---|---|---|---|---|
| | Ea(kJ/mol) | $R^2$ | $A(s^{-1})$ | Ea(kJ/mol) | $R^2$ | $A(s^{-1})$ | Ea(kJ/mol) | $R^2$ | $A(s^{-1})$ |
| 0.1 | 161.66 | 1.0000 | $2.83\times10^{14}$ | 198.63 | 0.9958 | $8.82\times10^{15}$ | 196.98 | 0.992 | $6.36\times10^{21}$ |
| 0.2 | 213.44 | 0.9951 | $1.10\times10^{19}$ | 197.61 | 0.9999 | $1.09\times10^{15}$ | 196.48 | 1.000 | $8.85\times10^{20}$ |
| 0.3 | 227.73 | 0.9973 | $9.82\times10^{19}$ | 212.34 | 0.9997 | $1.33\times10^{16}$ | 210.71 | 0.999 | $9.91\times10^{23}$ |
| 0.4 | 272.07 | 0.9944 | $3.56\times10^{23}$ | 231.19 | 0.9989 | $3.88\times10^{17}$ | 228.83 | 0.998 | $2.55\times10^{23}$ |
| 0.5 | 291.05 | 0.9923 | $2.74\times10^{24}$ | 312.34 | 0.9945 | $2.91\times10^{24}$ | 306.21 | 0.990 | $1.12\times10^{30}$ |
| 0.6 | 589.02 | 0.9207 | $7.09\times10^{48}$ | 404.39 | 0.9803 | $3.82\times10^{31}$ | 394.05 | 0.963 | $9.44\times10^{36}$ |
| 0.7 | 317.26 | 0.8376 | $1.33\times10^{24}$ | 437.53 | 0.9662 | $3.26\times10^{32}$ | 426.10 | 0.936 | $7.772\times10^{37}$ |
| Average | 296.03 | | - | 284.86 | | - | 279.91 | | - |

The kinetic analysis of SCG pyrolysis reveals distinct decomposition stages for hemicellulose, cellulose, and lignin, each with different activation energy requirements. Hemicellulose decomposes first at lower Ea values ($\alpha$=0.1 to 0.3), followed by cellulose at moderate Ea values ($\alpha$=0.4 to 0.6), and finally lignin, which requires higher Ea but decomposes more slowly over a broad temperature range ($\alpha$=0.7). The KAS model provides the most reliable results, with an average Ea of 284.86 kJ/mol and high $R^2$ values (>0.99) across most stages, making it the most consistent for describing SCG pyrolysis. While $R^2$ slightly declines at higher conversions ($R^2$ = 0.9662 at $\alpha$=0.7), this limitation is less pronounced than in the Friedman and FWO models.

The KAS model also offers physically realistic pre-exponential factors, aligning well with biomass reaction kinetics. In summary, SCG pyrolysis involves complex reactions across its components, with the KAS model emerging as the most effective for accurately estimating kinetic parameters.

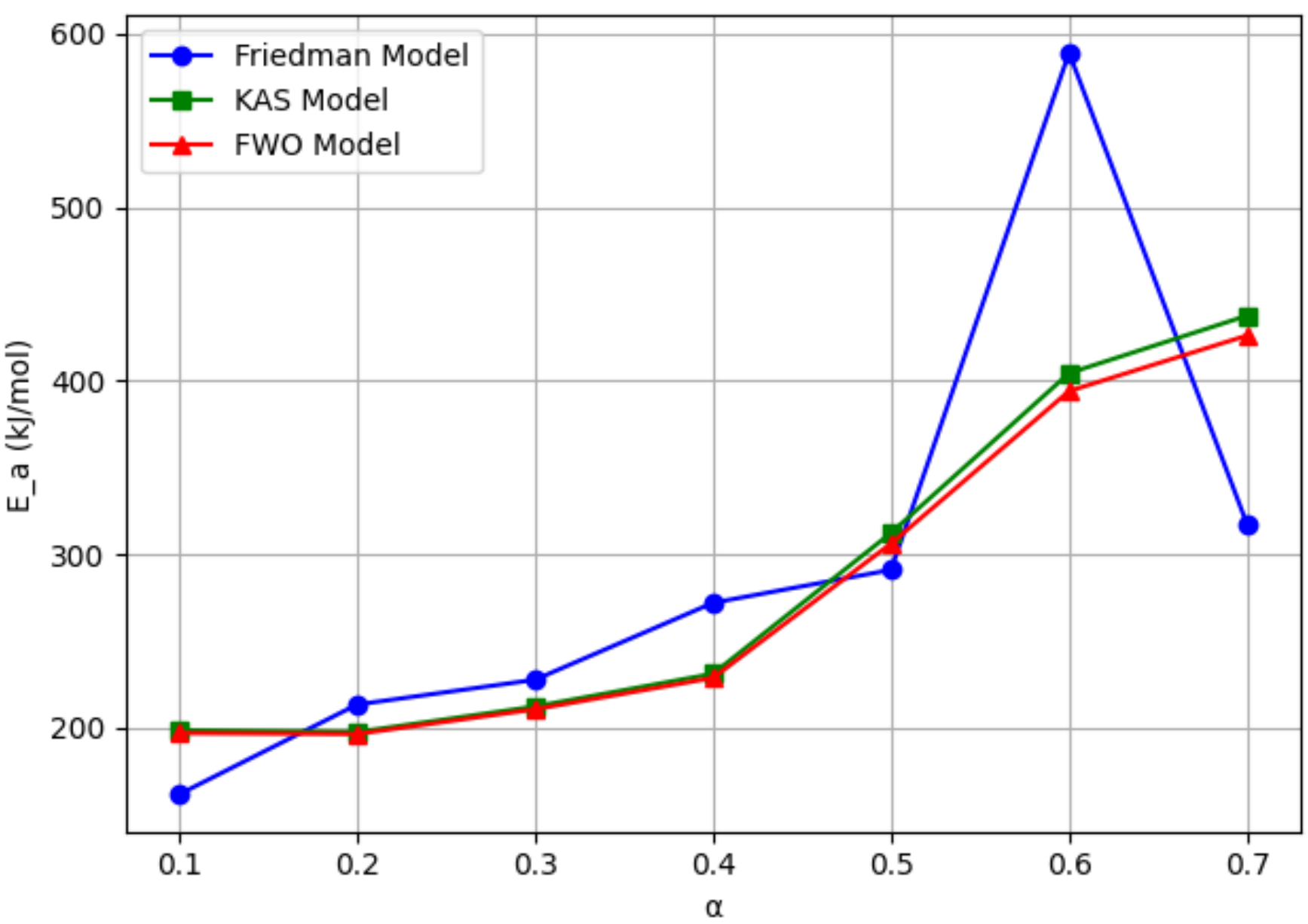

**Figure 10.** $E_a$ evolution as function of α for pure SCG (by Friedman, KAS and FWO models).

According to Table 6 and Figure 11, the thermal decomposition of date seeds (DS) involves distinct stages tied to the breakdown of its main components: hemicellulose, cellulose, and lignin. At lower conversions ($\alpha$ = 0.1 to $\alpha$ = 0.3), hemicellulose decomposition dominates. At lower conversions ($\alpha$=0.1–0.3), hemicellulose decomposes with moderate activation energy ($Ea$: 199.46–257.07 kJ/mol) and lower molecular activity (low A values). At intermediate conversions ($\alpha$=0.4–0.6), cellulose dominates, requiring higher Ea values (peaking at 409.81 kJ/mol in the Friedman model) and increased A values. At higher conversions ($\alpha$=0.7), lignin and residual materials decompose slowly, leading to a decline in Ea and $R^2$ values. This is noticed particularly in the KAS and FWO models, due to challenges in modelling these slow reactions. Among the models, the Friedman model shows high $R^2$ but unrealistic A values at higher conversions, suggesting overestimation. The FWO model is less reliable, with low $R^2$ and inconsistent Ea values at higher conversions. The KAS model strikes a balance, providing consistent Ea and realistic A values, making it the most reliable for this study. Overall, the thermal decomposition of DS progresses from hemicellulose to cellulose and concludes with lignin. Activation energy increases across these stages, reflecting the transition from less stable to more stable components.

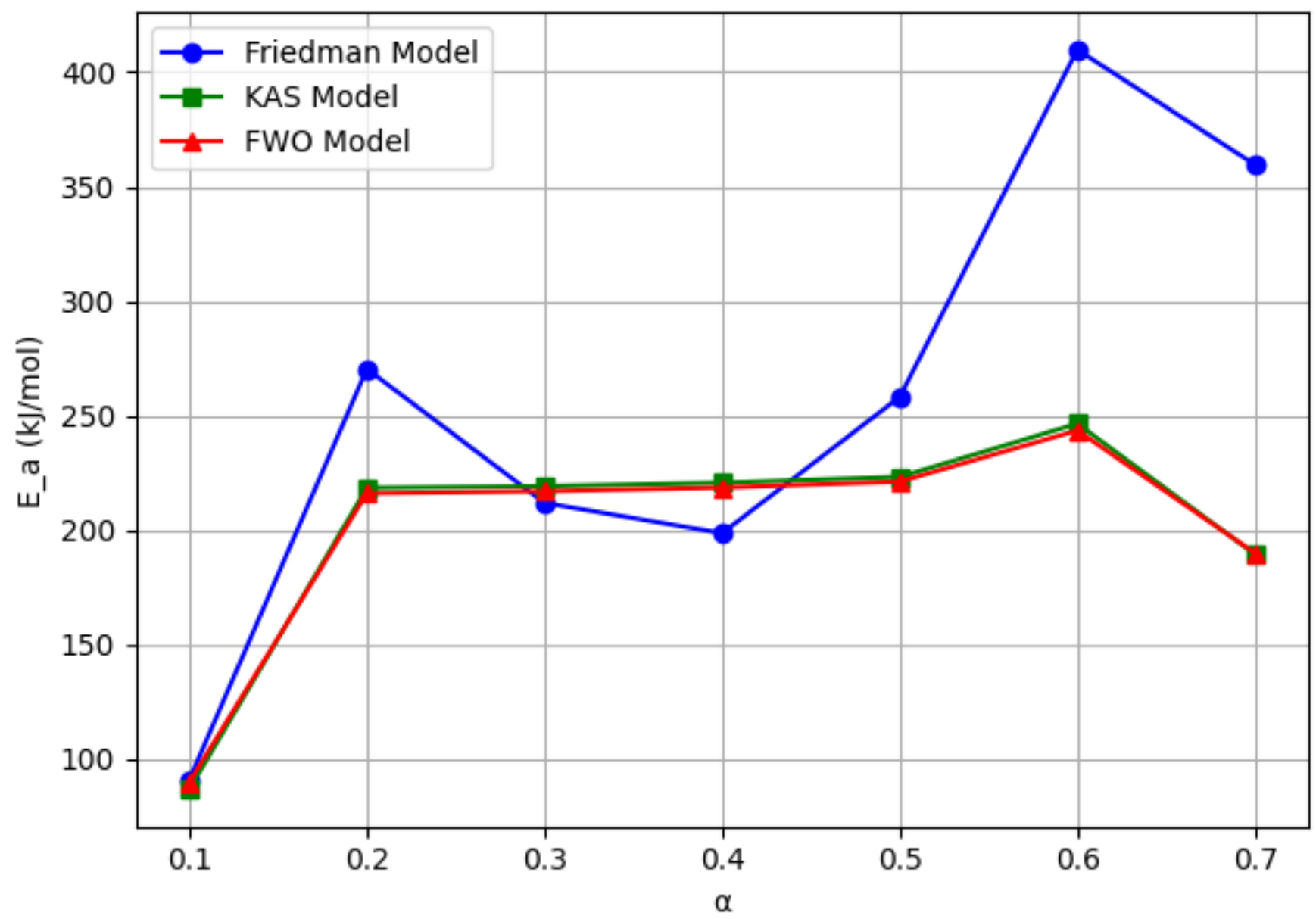

**Figure 11.** $E_a$ evolution as function of α for pure DS (by Friedman, KAS and FWO models).

**Table 6.** Kinetic Parameters from Friedman, KAS, & FWO Models for pure DS.

| α | Friedman Model | | | KAS Model | | | FWO Model | | |
|---|---|---|---|---|---|---|---|---|---|
| | Ea(kJ/mol) | $R^2$ | A($s^{-1}$) | Ea(kJ/mol) | $R^2$ | A($s^{-1}$) | Ea(kJ/mol) | $R^2$ | A($s^{-1}$) |
| 0.1 | 90.94 | 0.9945 | $1.11\times10^{8}$ | 86.78 | 0.9974 | $2.33\times10^{5}$ | 89.86 | 0.996 | $5.85\times10^{11}$ |
| 0.2 | 270.46 | 0.9997 | $1.53\times10^{25}$ | 218.55 | 0.9912 | $6.41\times10^{17}$ | 216.10 | 0.984 | $4.01\times10^{23}$ |
| 0.3 | 211.88 | 0.9964 | $1.06\times10^{19}$ | 219.18 | 0.9997 | $2.08\times10^{17}$ | 216.99 | 0.999 | $1.39\times10^{23}$ |
| 0.4 | 198.68 | 0.9886 | $3.71\times10^{17}$ | 220.72 | 0.9999 | $1.69\times10^{17}$ | 218.62 | 1.000 | $1.15\times10^{23}$ |
| 0.5 | 258.20 | 0.9964 | $4.06\times10^{22}$ | 223.28 | 0.9945 | $1.69\times10^{17}$ | 221.20 | 0.990 | $1.16\times10^{23}$ |
| 0.6 | 409.81 | 1.0000 | $1.75\times10^{35}$ | 246.52 | 0.9257 | $7.21\times10^{18}$ | 243.54 | 0.866 | $4.32\times10^{24}$ |
| 0.7 | 359.54 | 1.0000 | $3.81\times10^{28}$ | 189.47 | 0.7329 | $4.94\times10^{12}$ | 189.93 | 0.563 | $5.60\times10^{18}$ |
| Average | 257.07 | | - | 200.64 | | - | 199.46 | | - |

### 3.2.4. Kinetic analysis of blended Samples

The kinetic analysis of Blend 1 (75% DS, 25% SCG) (Figure 12 and Table 7) shows it decomposes more easily than its individual components, with significantly lower activation energy ($Ea$). The Friedman model provided the most accurate results, with an average Ea of 161.75 kJ/mol and high $R^2$ values, indicating a strong fit. In comparison, the KAS and FWO models produced lower and less consistent Ea and $R^2$ values. The reduced $Ea$ for Blend 1 suggests the blending of DS and SCG in these proportions enhances decomposition kinetics and improves energy efficiency.

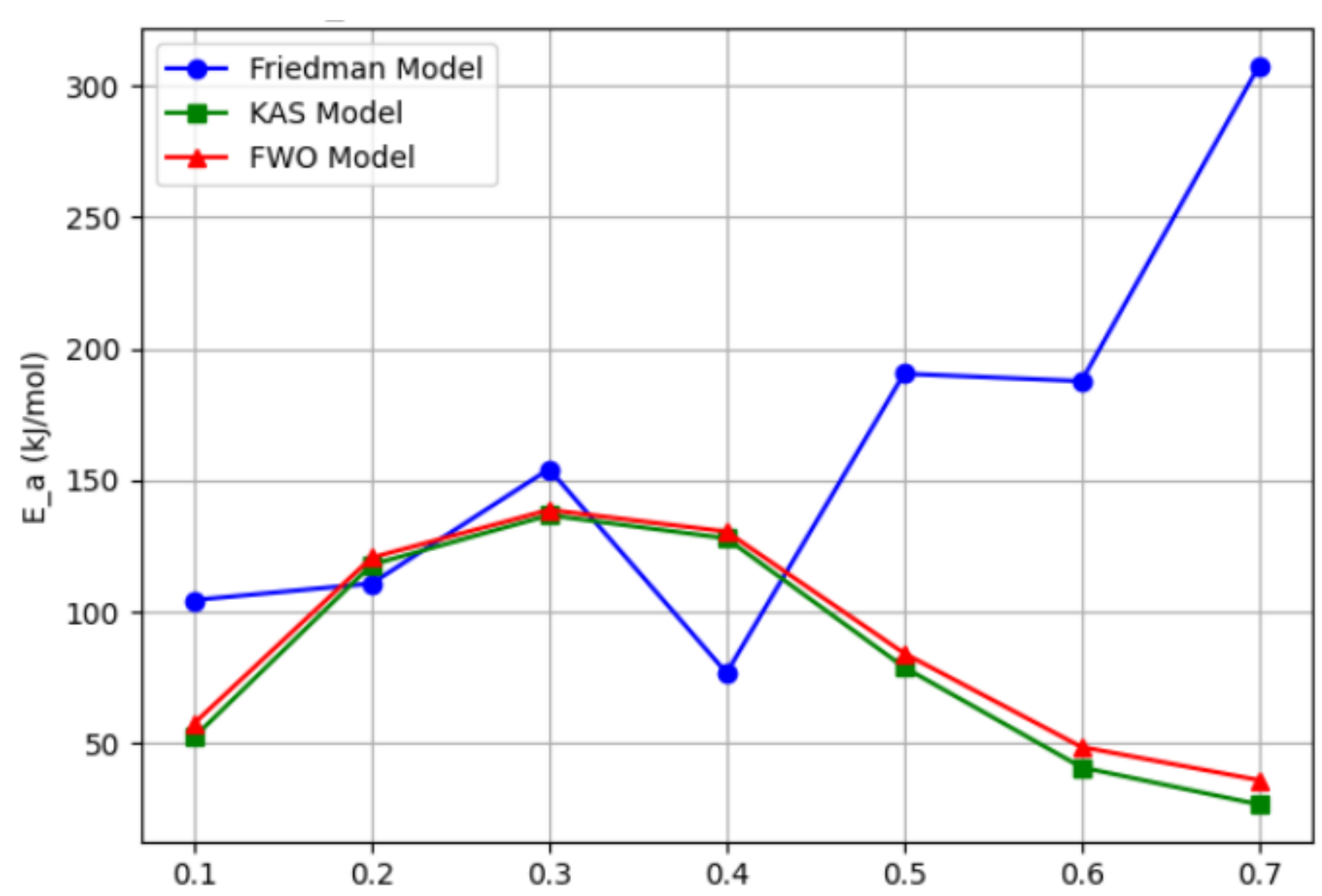

**Figure 12.** $E_a$ evolution as function of α for Blend 1 (by Friedman, KAS and FWO models).

**Table 7.** Comparison of Kinetic Parameters from Friedman, KAS, and FWO Models for 75% DS and 25% SCG.

| α | Friedman Model | | | KAS Model | | | FWO Model | | |
|---|---|---|---|---|---|---|---|---|---|
| | Ea(kJ/mol) | $R^2$ | $A(s^{-1})$ | Ea(kJ/mol) | $R^2$ | $A(s^{-1})$ | Ea(kJ/mol) | $R^2$ | $A(s^{-1})$ |
| 0.1 | 104.51 | 1.000 | $6.55\times10^{8}$ | 52.37 | 0.7285 | $9.26\times10^{0}$ | 57.61 | 0.602 | $7.95\times10^{7}$ |
| 0.2 | 110.76 | 0.9999 | $2.35\times10^{9}$ | 117.90 | 0.9705 | $2.51\times10^{7}$ | 120.61 | 0.950 | $5.44\times10^{13}$ |
| 0.3 | 154.30 | 0.9919 | $2.66\times10^{13}$ | 136.76 | 0.9896 | $1.30\times10^{9}$ | 138.76 | 0.982 | $2.23\times10^{15}$ |
| 0.4 | 76.79 | 0.9471 | $1.04\times10^{6}$ | 127.91 | 0.9871 | $1.48 \times10^{8}$ | 130.51 | 0.978 | $2.98\times10^{14}$ |
| 0.5 | 190.54 | 1.0000 | $1.15\times10^{16}$ | 78.95 | 0.9329 | $2.24\times10^{3}$ | 84.27 | 0.895 | $1.21\times10^{10}$ |
| 0.6 | 187.73 | 1.0000 | $8.53\times10^{14}$ | 40.88 | 0.8269 | $3.47\times10^{-1}$ | 48.62 | 0.773 | $7.04\times10^{6}$ |
| 0.7 | 307.60 | 1.0000 | $4.60\times10^{23}$ | 26.76 | 0.7409 | $1.10\times10^{-2}$ | 36.02 | 0.711 | $5.40\times10^{5}$ |
| Average | 161.75 | | - | 83.08 | | - | 88.06 | | - |

The kinetic analysis of Blend 2 (50% DS, 50% SCG) (Figure 13 and Table 8) shows notable differences between the models. The KAS model provides  stable activation energy (Ea) values averaging 292.49 kJ/mol and consistently high $R^2$ values above 0.92. The Friedman model The Friedman model also fits reasonably well, with an average $Ea$ of 284.01 kJ/mol, but shows significant variability, reducing its consistency. On the other hand, the FWO model provides much lower activation energies, starting at 14.95 kJ/mol at $\alpha = 0.1$ and peaking at 145.04 kJ/mol at $\alpha = 0.4$, with an average Ea of 80.91 kJ/mol. It is much lower than both the Friedman and KAS models, indicating that the FWO model may be underestimating the activation energy for Blend. Additionally, the $R^2$ values for the FWO model are much lower, particularly at higher $\alpha$ values, with the lowest being 0.047 at $\alpha = 0.7$. It suggests that the FWO model is not fitting the data well, especially at higher conversions.

Compared to pure SCG and DS, Blend 2 requires higher energy for pyrolysis, as indicated by its elevated $Ea$. This suggests that the 50/50 blend may be less energy-efficient for thermal decomposition than the individual components.

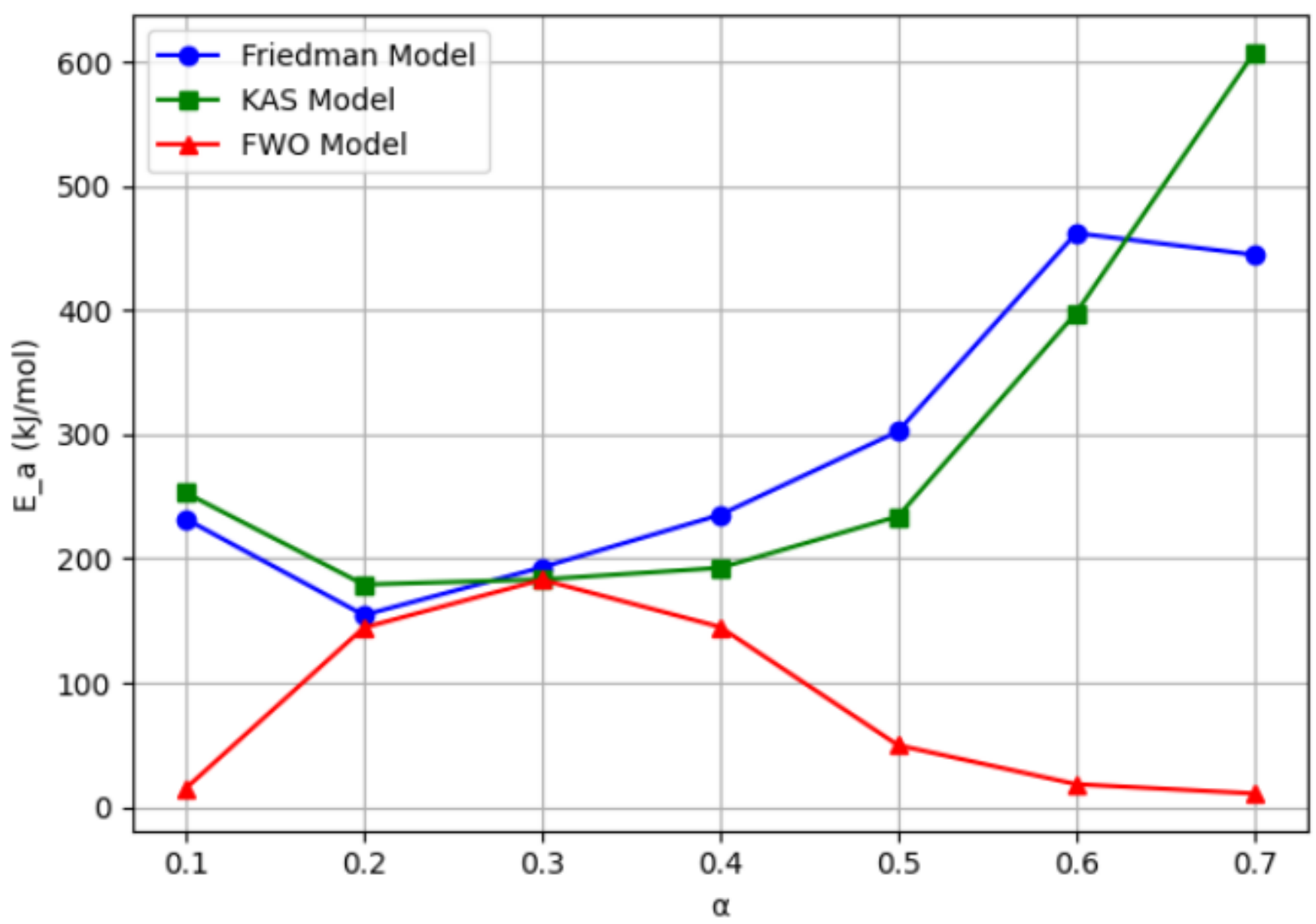

**Figure 13.** $E_a$ evolution as function of α for Blend 2 (by Friedman, KAS and FWO models).

**Table 8.** Comparison of Kinetic Parameters from Friedman, KAS, and FWO Models for 50% DS and 50% SCG.

| α | Friedman Model | | | KAS Model | | | FWO Model | | |
|---|---|---|---|---|---|---|---|---|---|
| | Ea(kJ/mol) | $R^2$ | A($s^{-1}$) | Ea(kJ/mol) | $R^2$ | A($s^{-1}$) | Ea(kJ/mol) | $R^2$ | A($s^{-1}$) |
| 0.1 | 231.98 | 0.8371 | $4.45\times10^{22}$ | 253.01 | 0.9270 | $1.81\times10^{23}$ | 14.95 | 0.093 | $7.95\times10^{3}$ |
| 0.2 | 154.57 | 0.8146 | $3.49\times10^{13}$ | 179.16 | 0.9247 | $2.93\times10^{13}$ | 144.43 | 0.726 | $5.44\times10^{16}$ |
| 0.3 | 193.00 | 0.8675 | $1.08\times10^{17}$ | 183.28 | 0.9247 | $3.77\times10^{13}$ | 183.01 | 0.867 | $2.23\times10^{19}$ |
| 0.4 | 235.26 | 0.8983 | $4.06\times10^{20}$ | 192.56 | 0.9249 | $1.85\times10^{14}$ | 145.04 | 0.686 | $5.91\times10^{15}$ |
| 0.5 | 302.45 | 0.9094 | $9.76\times10^{25}$ | 234.02 | 0.9259 | $6.44\times10^{17}$ | 49.75 | 0.240 | $1.01\times10^{7}$ |
| 0.6 | 462.11 | 0.8931 | $4.43\times10^{38}$ | 397.92 | 0.9279 | $3.92\times10^{31}$ | 18.37 | 0.080 | $3.41\times10^{4}$ |
| 0.7 | 444.73 | 0.7712 | $4.86\times10^{34}$ | 607.49 | 0.9289 | $8.43\times10^{46}$ | 10.85 | 0.047 | $1.39\times10^{4}$ |
| Average | 284.01 | | - | 292.49 | | - | 80.91 | | - |

The kinetic analysis of Blend 3 (25% DS, 75% SCG) (Figure.14 and Table 9) reveals significant differences in the models' performance. The Friedman model stands out as the most reliable, with variability in $Ea$ (108.59–566.78 kJ/mol) and fluctuating $R^2$.

The KAS model provides more consistent Ea values, ranging from 53.60 kJ/mol at $\alpha = 0.1$ to 649.81 kJ/mol at $\alpha = 0.7$, with an average Ea of 331.24 kJ/mol. The $R^2$ values for the KAS model are generally high, with excellent fit observed at $\alpha = 0.7$ ($R^2 = 0.999$), but there are deviations at lower ($R^2 = 0.233$ at $\alpha = 0.1$. The A values for the KAS model also show significant variability, reaching up to $1.43 \times 10^{50}$ $s^{-1}$ at $\alpha = 0.7$, indicating a high frequency of collisions between the components of the blend.

The FWO model closely mirrors the trends of the KAS model, with Ea values ranging from 58.90 kJ/mol at $\alpha$ = 0.1 to 627.80 kJ/mol at $\alpha$ = 0.7, and an average Ea of 295.31 kJ/mol. However, the $R^2$ values for the FWO model are also inconsistent at lower conversions ($R^2$ = 0.071 at $\alpha$ = 0.1), despite excellent fits at higher conversions ($R^2$ = 0.999 at $\alpha$ = 0.7). Additionally, the A values for the FWO model exhibit extreme variations, reaching $1.53 \times 10^{55}$ $s^{-1}$ at $\alpha$ = 0.7.

Overall, the average Ea values across the three models are relatively close, with Friedman (325.22 kJ/mol), KAS (331.24 kJ/mol), and FWO (295.31 kJ/mol) providing similar energy approximations. Blending DS and SCG increases the pyrolysis energy requirement compared to the pure components. The blend's average $Ea$ is higher than SCG (284.86 kJ/mol) and DS (200.64 kJ/mol), suggesting antagonistic interactions that make thermal degradation more complex.

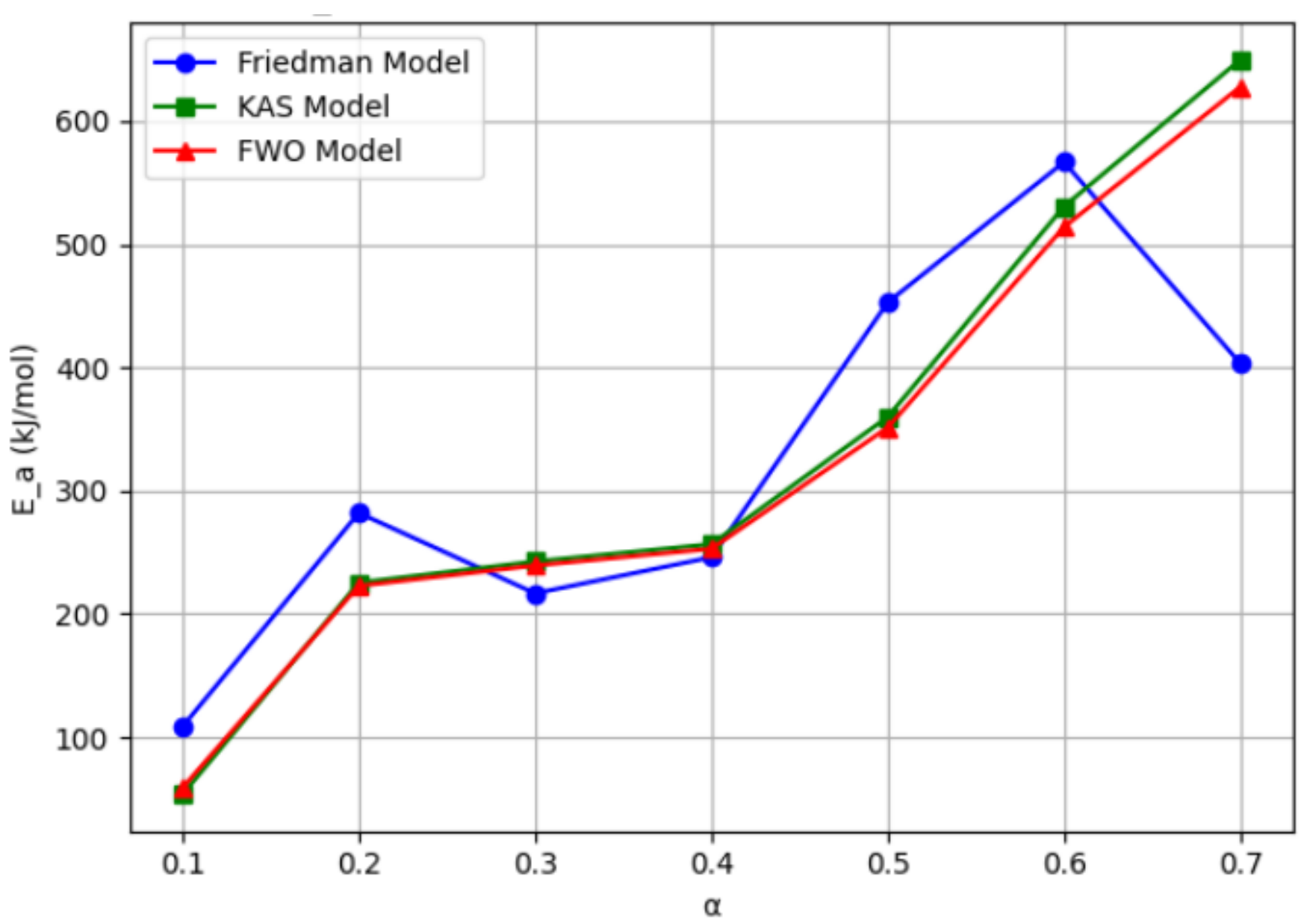

**Figure 14.** $E_a$ evolution as function of α for Blend 3 (by Friedman, KAS and FWO models).

Table 9. Kinetic Parameters from Friedman, KAS, and FWO Models for 25% DS and 75% SCG.

| α | Friedman Model | | | KAS Model | | | FWO Model | | |
|---|---|---|---|---|---|---|---|---|---|
| | Ea(kJ/mol) | $R^2$ | A($s^{-1}$) | Ea(kJ/mol) | $R^2$ | A($s^{-1}$) | Ea(kJ/mol) | $R^2$ | A($s^{-1}$) |
| 0.1 | 108.59 | 0.2954 | $1.39\times10^{9}$ | 53.60 | 0.233 | $9.45\times10^{0}$ | 58.90 | 0.071 | $7.98\times10^{7}$ |
| 0.2 | 282.07 | 0.9904 | $5.60\times10^{25}$ | 225.43 | 0.982 | $8.66\times10^{17}$ | 222.84 | 0.966 | $5.34\times10^{23}$ |
| 0.3 | 616.75 | 0.9530 | $1.53\times10^{19}$ | 242.59 | 0.990 | $1.55\times10^{19}$ | 239.39 | 0.982 | $8.71\times10^{24}$ |
| 0.4 | 246.01 | 0.9909 | $2.93\times10^{21}$ | 256.65 | 0.990 | $1.53 \times10^{20}$ | 252.93 | 0.982 | $8.02\times10^{25}$ |
| 0.5 | 452.89 | 0.9821 | $1.79\times10^{39}$ | 359.88 | 0.989 | $1.28\times10^{29}$ | 351.28 | 0.979 | $3.63\times10^{34}$ |
| 0.6 | 566.78 | 0.9934 | $2.92\times10^{47}$ | 530.73 | 0.978 | $8.53\times10^{42}$ | 514.03 | 0.957 | $1.21\times10^{48}$ |
| 0.7 | 403.44 | 0.9798 | $1.58\times10^{31}$ | 649.81 | 0.999 | $1.43\times10^{50}$ | 627.80 | 0.999 | $1.53\times10^{55}$ |
| Average | 325.22 | | - | 331.24 | | - | 295.31 | | - |

The analysis of the kinetic parameters for the three blends reveals that the blend ratio can significantly affect the activation energy (Ea) and the energy efficiency of the pyrolysis process. Among the blends, Blend 1 emerges as the most energy-efficient option compared to the other blends and pure samples, with the lowest average activation energy. The moderate variability in the pre-exponential factor (A) for Blend 1 further suggests a less complex reaction mechanism, making it the best choice for energy-efficient pyrolysis compared to the other blends and pure samples. In contrast, Blend 2 exhibits much higher Ea values, which indicates that it requires significantly higher energy for pyrolysis, likely due to increased complexity from interactions between SCG and DS components, making it less energy efficient. Similarly, Blend 3 also shows high Ea values across all models. This high energy demand suggests that Blend 3 is the least energy-efficient. It can be concluded that reducing the proportion of SCG in the blend decreases the activation energy, indicating that incorporating DS into SCG improves the efficiency of the pyrolysis process.

## 3.3. Thermodynamic Study

### 3.3.1. Pure samples

The thermal decomposition of SCG occurs in distinct phases linked to the degradation of hemicellulose, cellulose, and lignin. In Figure 15 and during the early stages ($\alpha$=0.1–0.4), moderate enthalpy ($\Delta H$) and stable Gibbs free energy ($\Delta G$) reflect the energy needed for initial bond cleavage. The mid stage ($\alpha$=0.5–0.6) is the most energy-intensive, particularly in the Friedman model, as lignin's stable structures are degraded. At later stages ($\alpha$=0.7), energy requirements and entropy ($\Delta S$) decline, likely due to the formation of stable char residues. This behaviour aligns with findings by Alsulami et al.[34] , who observed rising $\Delta G$ during lignin decomposition, peaking at $\alpha$=0.7 before decreasing. The increasing $\Delta G$ highlights SCG's potential for energy production but also underscores the challenges of efficient pyrolysis, as noted in previous studies on coffee waste [35] and maize cob biomass[36]. Variations in entropy further illustrate changes in molecular randomness during thermal degradation, consistent with trends observed in other biomass systems[37].

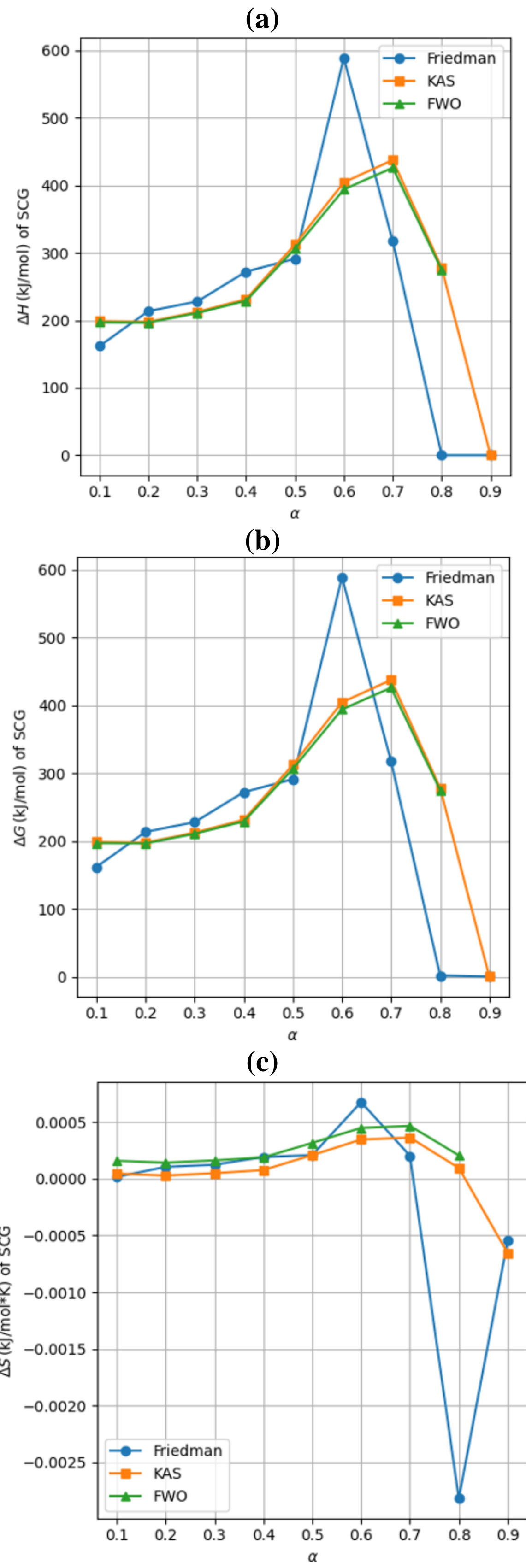

**Figure 15.** Pure SCG thermodynamic parameters (by Friedman, KAS, & FWO Models): (a) ΔH; (b) ΔG; (c) ΔS.

According to Figure 16, the increasing values of $\Delta H$ and $\Delta G$ during the intermediate stages of pyrolysis emphasize the energy-intensive breakdown of cellulose and lignin.

Conversely, the decreasing $\Delta S$ values at higher conversion rates suggest a shift toward char formation and structural stabilization. Notably, the entropy change further supports the idea of increasing disorder at early stages of decomposition, stabilizing as the process reaches higher conversions, and indicating the complexity of the thermal degradation process in date seeds.

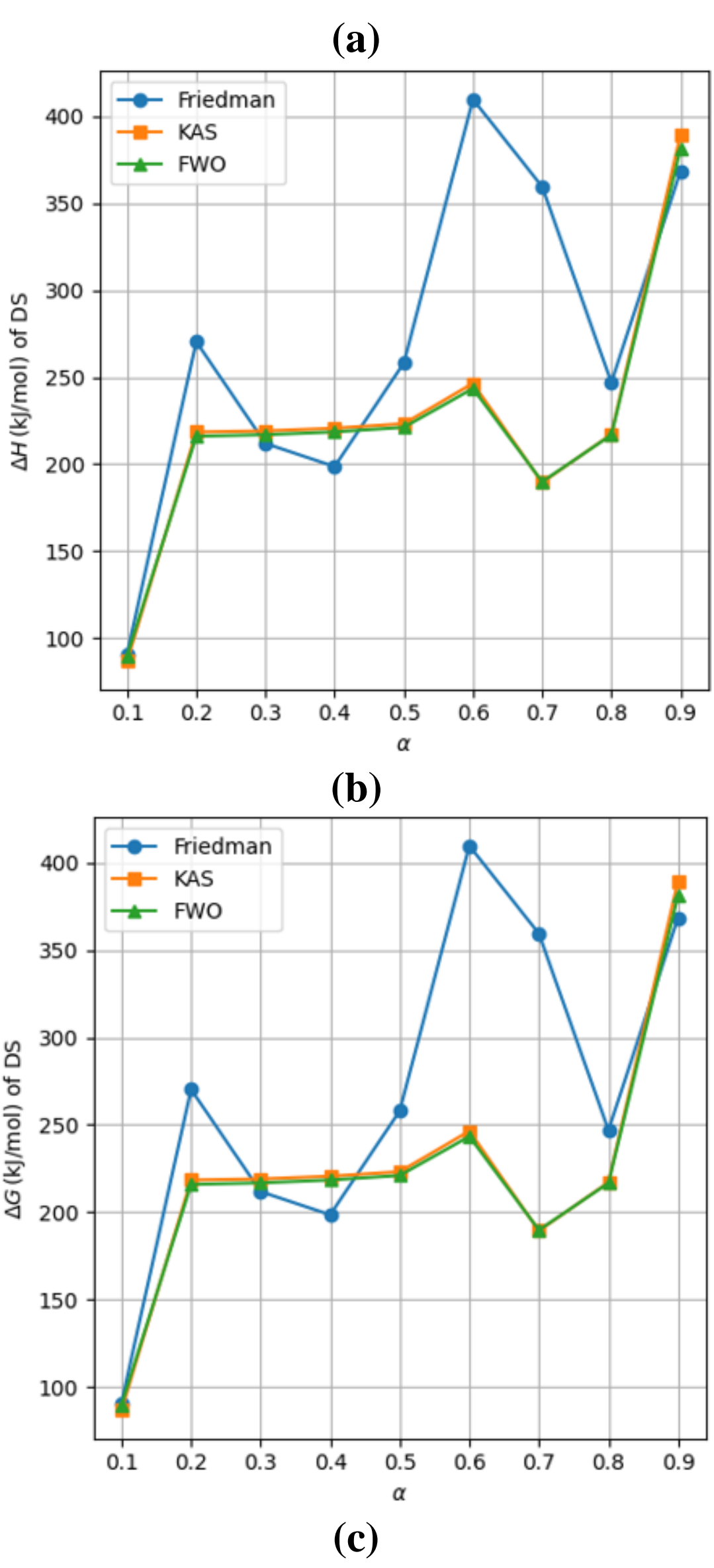

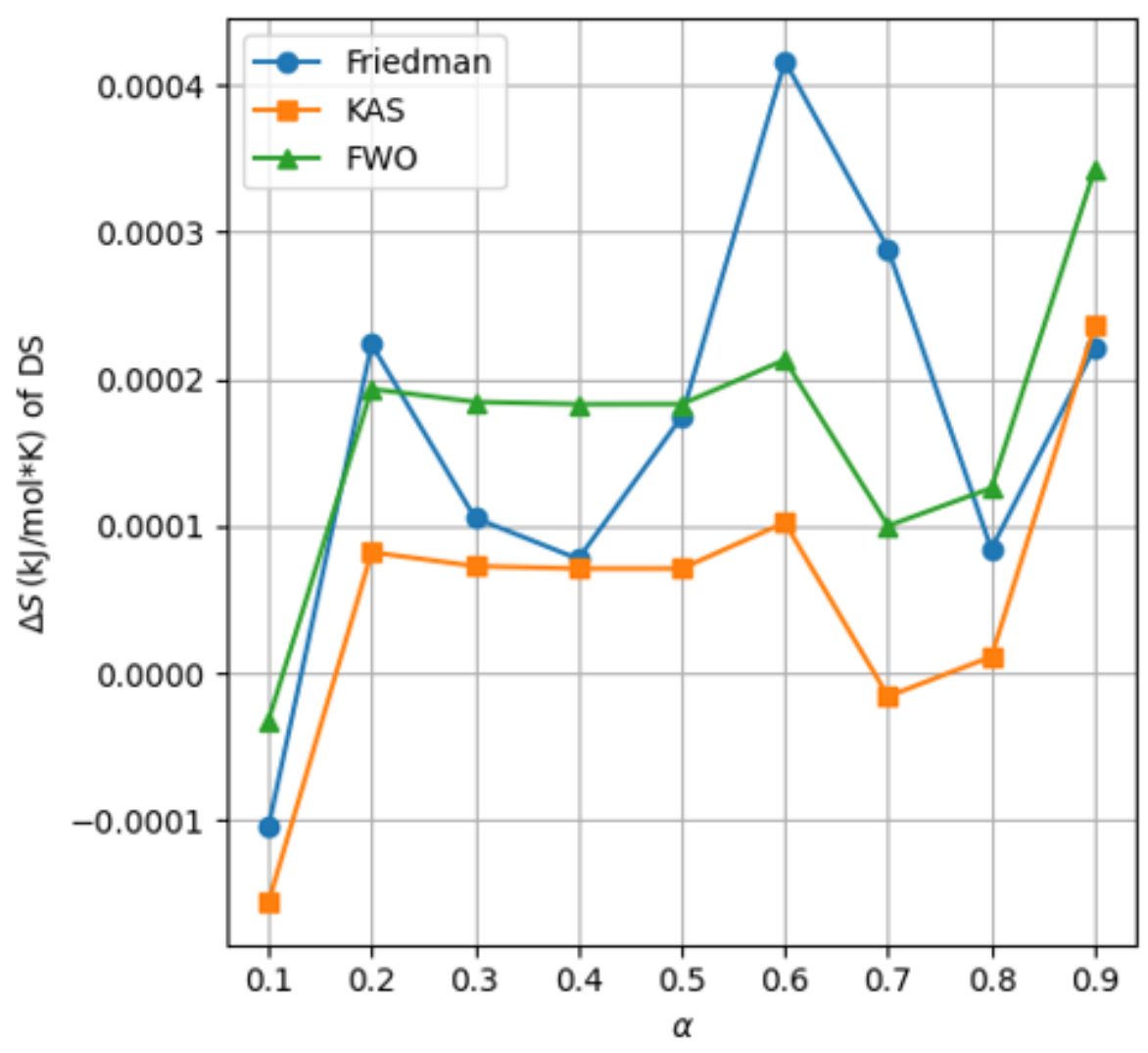

**Figure 16.** Pure DS thermodynamic parameters (by Friedman, KAS, & FWO Models): (a) ΔH; (b) ΔG; (c) ΔS.

### 3.3.2. Blended samples

According to Figure 17, the Friedman model provides consistent thermodynamic values, with the high average enthalpy indicating a significant energy requirement, which is typical for biomass decomposition. The Gibbs free energy follows a similar trend as enthalpy, remaining positive and suggesting a non-spontaneous process that requires external energy input. The entropy change is modest, reflecting a moderately ordered transition during decomposition. Additionally, the averaged thermodynamic values for Blend 1 are lower than those of the pure samples, supporting the earlier conclusion that the higher proportion of DS in the blend leads to a more energy-efficient pyrolysis process compared to the individual components. This further confirms that the addition of DS reduces the overall energy demand.

**(a)**

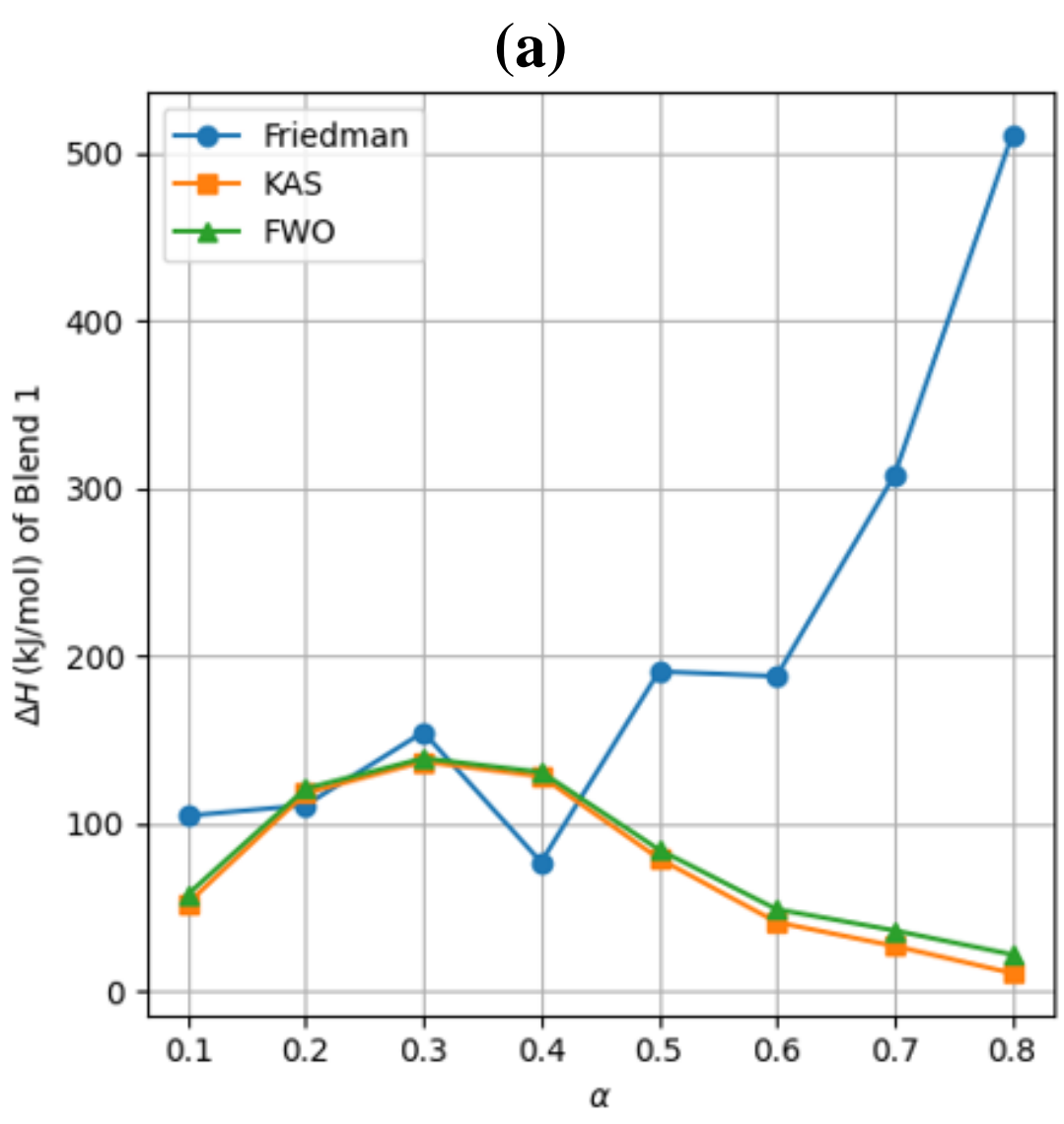

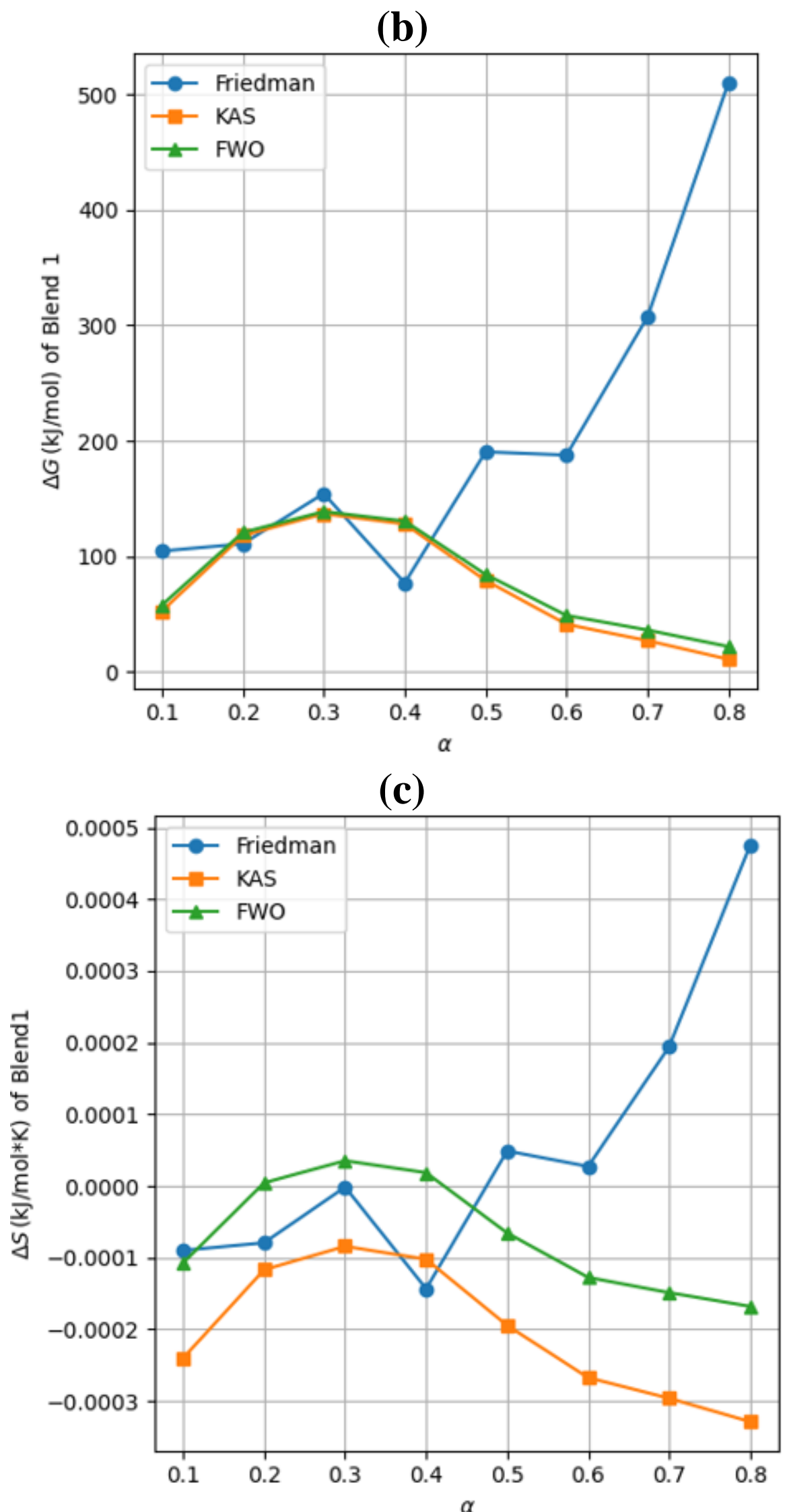

**Figure 17.** Blends 1 thermodynamic parameters (by Friedman, KAS, & FWO Models): (a) ΔH; (b) ΔG; (c) ΔS.

The decomposition of Blend 2 (50% DS, 50% SCG) involves complex thermodynamic phases and the enthalpy and Gibbs free energy follow the same trend as the DS sample and Blend 1 in the initial and mid stages (Figure 18). The late stages ($\alpha$=0.7) show a decline in energy requirements and a more stable entropy profile, which may reflect char formation and the stabilization of the remaining biomass. The KAS model, with its smoother trends, seems to offer the most reliable representation of thermal behaviour for this blend, while the Friedman model's sharp fluctuations may be attributed to its sensitivity to experimental conditions.

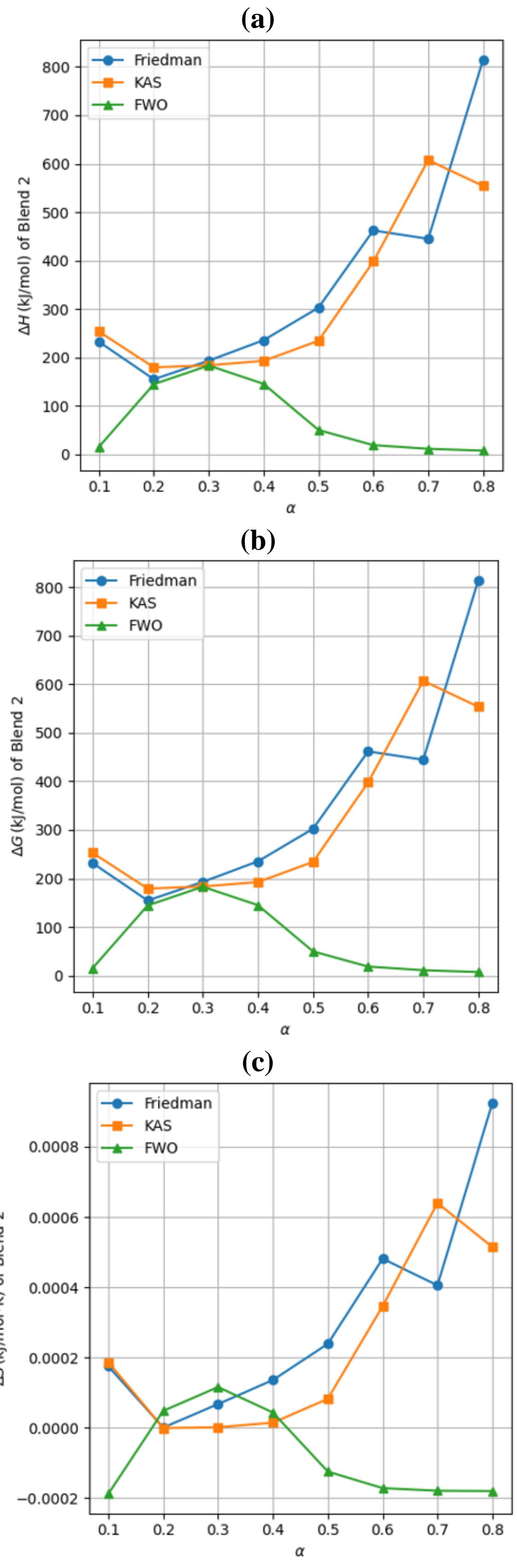

**Figure 18.** Blends 2 thermodynamic parameters (by Friedman, KAS, & FWO Models): (a) ΔH; (b) ΔG; (c) ΔS.

The decomposition of Blend 3 (25% DS, 75% SCG) exhibits distinct thermodynamic phases, similar to the other biomass blends (Figure 19). At the late stage ($\alpha$=0.7), energy requirements and entropy decline. The KAS model, with its stable thermodynamic parameters, proves to be the most robust method for analyzing both kinetic and thermodynamic behaviour during Blend 3 decomposition, while the Friedman model, although providing more detailed insights, is highly sensitive to experimental variations. Overall, the thermodynamic parameters indicate that Blend 3 requires considerable external energy for pyrolysis, with a higher energy demand than all of the other samples, reflecting the higher proportion of SCG in this blend.

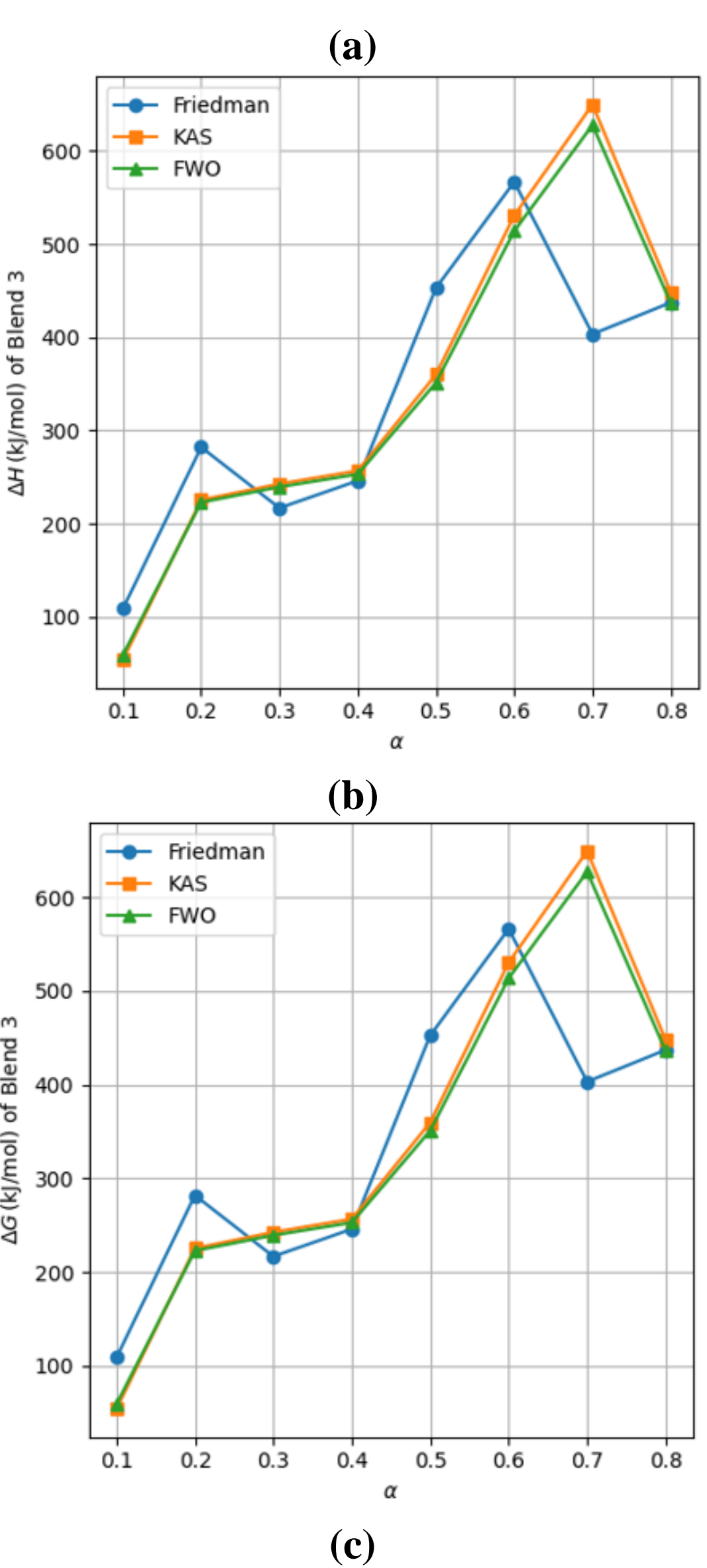

(c)

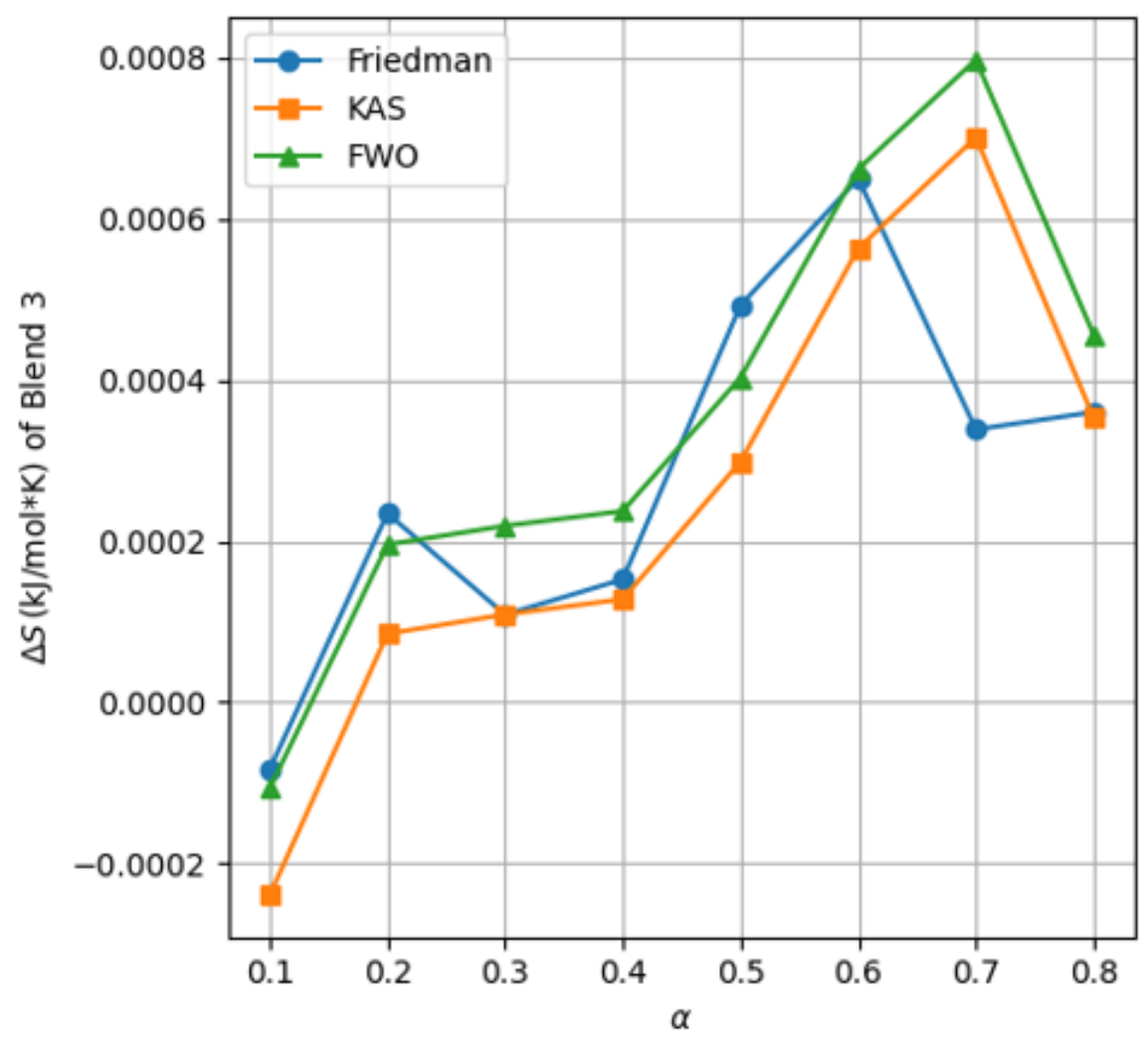

**Figure 19.** Blends 3 thermodynamic parameters (by Friedman, KAS, & FWO Models): (a) ΔH; (b) ΔG; (c) ΔS.

## 3.4. Pyrolysis product analysis

### 3.4.1. Solid yields (char & volatiles)

The pyrolysis reaction was performed on the five different samples, as given above. The results for each test are summarized in Table 10. The η values represent the char percentage yielded after the pyrolysis process, and the VM% is the volatile matter yielded by each sample and calculated according to the equation (1):

$$VM\% = 100\% - \eta(\%) \qquad (1)$$

The pyrolysis tests results indicate a clear trend of increasing volatile matter (VM) and decreasing char yield ($\eta$) with higher SCG proportions in the blends. Pure DS yields 72.26% VM and 27.74% char, while pure SCG has slightly lower VM (72.16%) and slightly higher char yield (27.84%). The moderate increase in volatile matter for the 50% DS - 50% SCG blend (71.59%) compared to pure DS (72.26%) indicates a synergistic effect where SCG promotes volatile release when mixed with DS. Blend 3 (25% DS, 75% SCG) exhibits the highest VM (73.61 %) and lowest char yield (26.39 %), highlighting SCG's role in promoting volatile release. These findings demonstrate that blending DS with SCG alters thermal degradation dynamics, favouring volatile release over char production.

**Table 10.** Char Yield and Volatile Matter Percentage for Different Biomass Samples.

| Sample | VM yield % | Char yield (η) % |
|---|---|---|
| 100 % DS | 72.26 | 27.74 |
| 100 % SCG | 72.16 | 27.84 |
| 75 % DS – 25 % SCG | 70.69 | 29.04 |
| 50 % DS – 50 % SCG | 71.59 | 28.41 |
| 25 % DS – 75 % SCG | 73.61 | 26.39 |

Pyrolysis involves heating a sample under controlled conditions, breaking it into smaller fragments, which are then analysed using Micro-GC to identify and quantify the resulting components. In our specific case, the gases were collected in a Tedlar bag after pyrolysis of biomass at a heating rate of 10°C/min and a final temperature of 600°C and subsequently injected into the Micro-GC for analysis. In this study, the Py-micro-GC analysis provided data on various gas components, including their concentrations expressed as mole percentages and parts per million (ppm) concentrations. The measured gases included hydrogen ($H_2$), methane ($CH_4$), nitrogen ($N_2$), acetylene ($C_2H_2$), oxygen ($O_2$), carbon monoxide (CO), and carbon dioxide ($CO_2$). Ethane ($C_2H_6$) was also quantified to further characterize the pyrolysis gas.

### 3.4.2. Incondensable gas (Micro-GC)

The Micro GC analysis reveals notable differences in the gas production during pyrolysis under specific conditions (600°C as a final temperature and 10°C/min as a heating rate) of DS and SCG (Table 11), with SCG generally releasing higher concentrations of hydrogen and light hydrocarbons compared to DS. Specifically, SCG produces 4.6259 mole % of hydrogen, significantly more than the 0.9024 mole % from DS, suggesting SCG is a better candidate for processes such as hydrogen fuel production. Additionally, DS yields more methane (2.4889 mole %) than SCG, making it more suited for chemical synthesis or energy production. SCG also generates more ethylene and ethane, further emphasizing its potential in chemical processing and approaches that can capture hydrogen from these components, facilitating the production of additional $H_2$ through processes such as catalytic reforming. However, DS could still be valuable in methane recovery applications or other lower-energy applications, despite its lower energy content.

**Table 11.** Gas Analysis for DS and SCG samples pyrolysis.

| Component | Concentration in mole (%) | |
|---|---|---|
| | DS | SCG |
| $H_2$ | 0.9024 | 4.6259 |
| $O_2$ | - | - |
| $CH_4$ | 2.4889 | 1.0785 |
| CO | - | - |
| $N_2$ | 80.4890 | 67.0377 |
| $O_2$ | - | 13.5678 |
| $C_2H_2$+$C_2H_4$ | 0.1235 | 0.4650 |
| $C_2H_6$ | 0.1901 | 0.6095 |

The gas analysis for Blends 1, 2, and 3 (Table 12), which consist of varying proportions of DS and SCG, reveals notable differences in the gaseous components produced during the previously mentioned pyrolysis conditions. As the proportion of SCG increases, so do the concentrations of hydrogen ($H_2$, methane ($CH_4$), ethylene + ethyne ($C_2H_2$ + $C_2H_4$), and ethane ($C_2H_6$), with Blend 3 (75 % SCG) exhibiting the highest values for all these gases, which aligns with the results of Table 10. In contrast, Blend 2 (50 % DS, 50 % SCG) shows moderate $H_2$ and other hydrocarbon yields compared to Blend 3. Blend 1 (75 % DS) produces fewer valuable gases compared to the other two blends. Therefore, Blend 3 stands out as the most promising blend for hydrogen generation, aligning with the results presented in the previous section.

**Table 12.** Gas Analysis for blends 1, 2 & 3.

| Component | Concentration in mole (%) | | |
|---|---|---|---|
| | Blend 1* | Blend 2* | Blend 3* |
| $H_2$ | 2.0277 | 2.7050 | 3.4378 |
| $O_2$ | - | - | - |
| $CH_4$ | 0.9947 | 1.3487 | 1.5960 |
| CO | - | - | - |
| $N_2$ | 68.2472 | 77.5777 | 70.0218 |
| $O_2$ | - | - | - |
| $C_2H_2$+$C_2H_4$ | 0.2786 | 0.4604 | 0.5598 |
| $C_2H_6$ | 0.3947 | 0.5901 | 0.6186 |

*Blend1: 75%DS25%SCG, Blend2:50%DS50%SCG, Blend3:25%DS 75%SCG

By utilizing agricultural waste such as date pits and coffee grounds in pyrolysis and co-pyrolysis processes, we can effectively convert these materials into valuable products like bio-oil, biogas, and biochar. This approach not only helps in the efficient utilization of agricultural waste but also contributes to reducing dependency on fossil fuels and mitigating environmental pollution caused by improper waste management.

The results of this study's experimental sections may serve as a helpful manual for producing biogas and biochar on an industrial scale from various agricultural wastes. The findings of the study provide useful guidance for the industrial-scale production of H2 and biochar from these feedstocks, which can be used as a catalytic support, water pollutant adsorbent, and gas adsorbent in various chemical production, purification, and storage processes. The synthesis of biohydrogen from waste lignocellulosic biomass accomplishes the dual purpose of converting waste into valuable products and reduces waste disposal issues. Co-pyrolysis has been identified as a promising method for improving the performance of the biomass pyrolysis process through synergistic interactions.

### 3.5. Prediction results

This subsection presents the obtained results of the LSTM model with two different datasets for pure biomasses and their blends. Model 1, which only uses basic features (temperature, heating rate, and blend ratios), achieving comparable or superior accuracy to ANN models from the literature. For instance, Tariq et al. [38] achieved $R^2$ values of 0.975–0.996 for TGA predictions of biomass blends, while Model 1 in this study yielded similar $R^2$ values (0.9658–0.9958) (Table 13), despite using a smaller, less diverse dataset. Another relevant study by Yildiz et al. [39] reported an exceptional $R^2$ of 0.9995 for predicting TGA curves of hazelnut husk and lignite coal blends at varying heating rates. Their results slightly outperform Model 1 (Table 13), likely due to their dataset's greater diversity in heating rates and blend ratios. Similarly, Bi et al. [40] achieved an $R^2$ of 0.99972 by incorporating a wide range of blend ratios and heating rates in their ANN model. These findings emphasize the importance of dataset diversification to enhance model generalization.

Model 1's limitations remain, particularly with predicting profiles for higher heating rates and unseen blend ratios, underscoring the need for more diverse training data (Figures 20 and 21). Despite these differences, the LSTM models exhibited robust performance, as demonstrated by the achieved loss metrics (Table 13).

**Table 13.** Comparison of Error Metrics: Model 1 vs. Model2.

| **Error Metrics** | **Model 1*** | | | **Model 2*** | | |
|---|---|---|---|---|---|---|
| | 10°C/min | 15°C/min | 25°C/min | 10°C/min | 15°C/min | 25°C/min |
| **$R^2$** | 0.9958 | 0.9658 | 0.9830 | 0.9996 | 0.9996 | 0.9998 |
| **MSE** | 3.6557 | 39.3286 | 14.9231 | 0.3586 | 0.4637 | 0.1809 |
| **MAE** | 1.6525 | 4.6805 | 3.1518 | 0.4206 | 0.4828 | 0.3509 |
| **RMSE** | 1.9120 | 6.2713 | 3.8630 | 0.5988 | 0.6809 | 0.4254 |

*Model 1: Trained on temperature, heating rates and ratios
*Model 2: trained on lignocellulosic composition, temperature, heating rates and ratio features

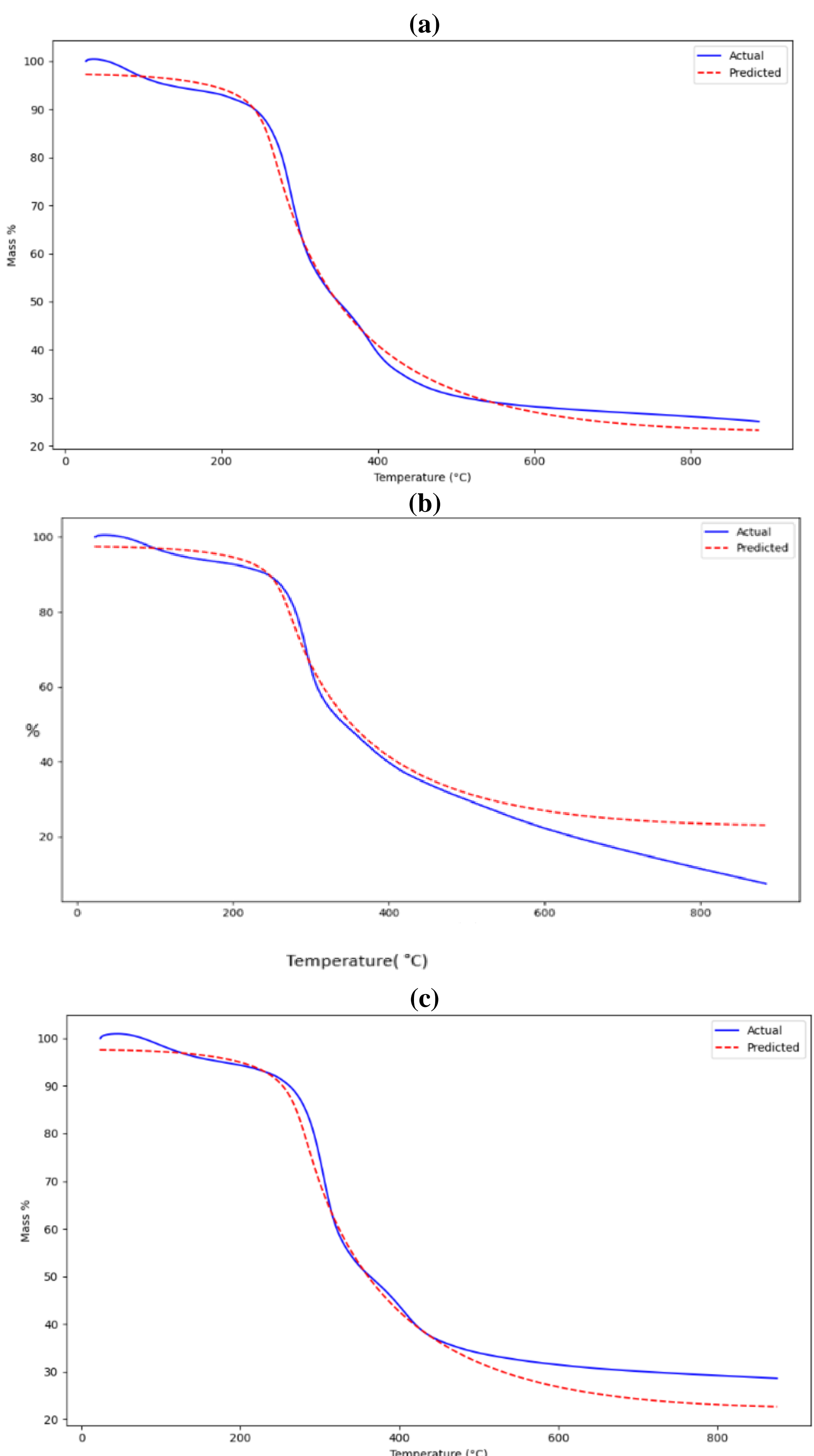

**Figure 20.** Model 1's Prediction of Blend 1's TGA data: (a) 10°C/min; (b) 15°C/min; (c) 25°C/min.

Incorporating lignocellulosic composition as additional features in Model 2 further enhanced prediction accuracy, especially for unseen heating rates like 25°C/min, as shown in Figures 21a, b and c. This improvement is evident in the significantly lower error metrics (MSE and MAE) and higher $R^2$ values compared to Model 1 (Table 13).

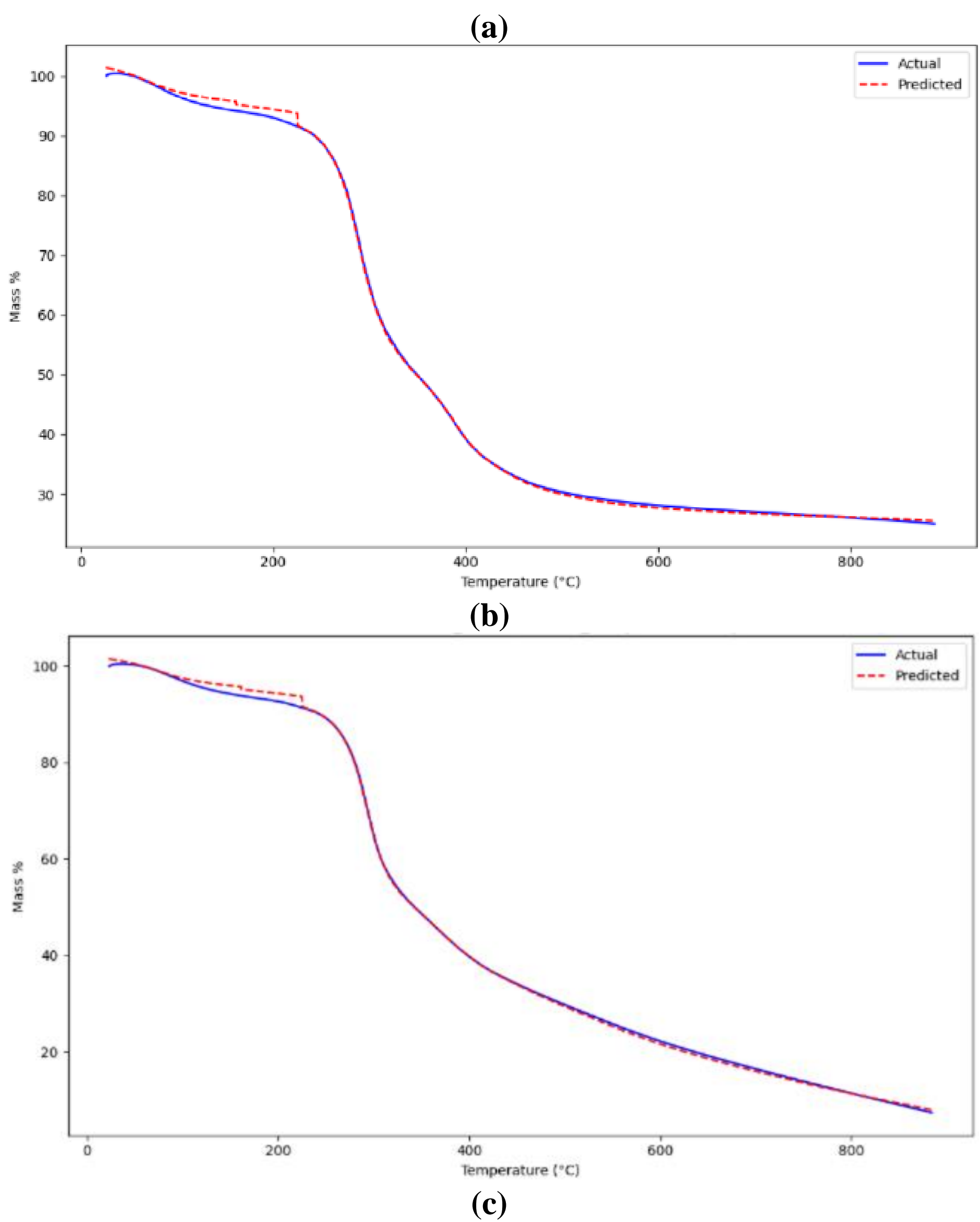

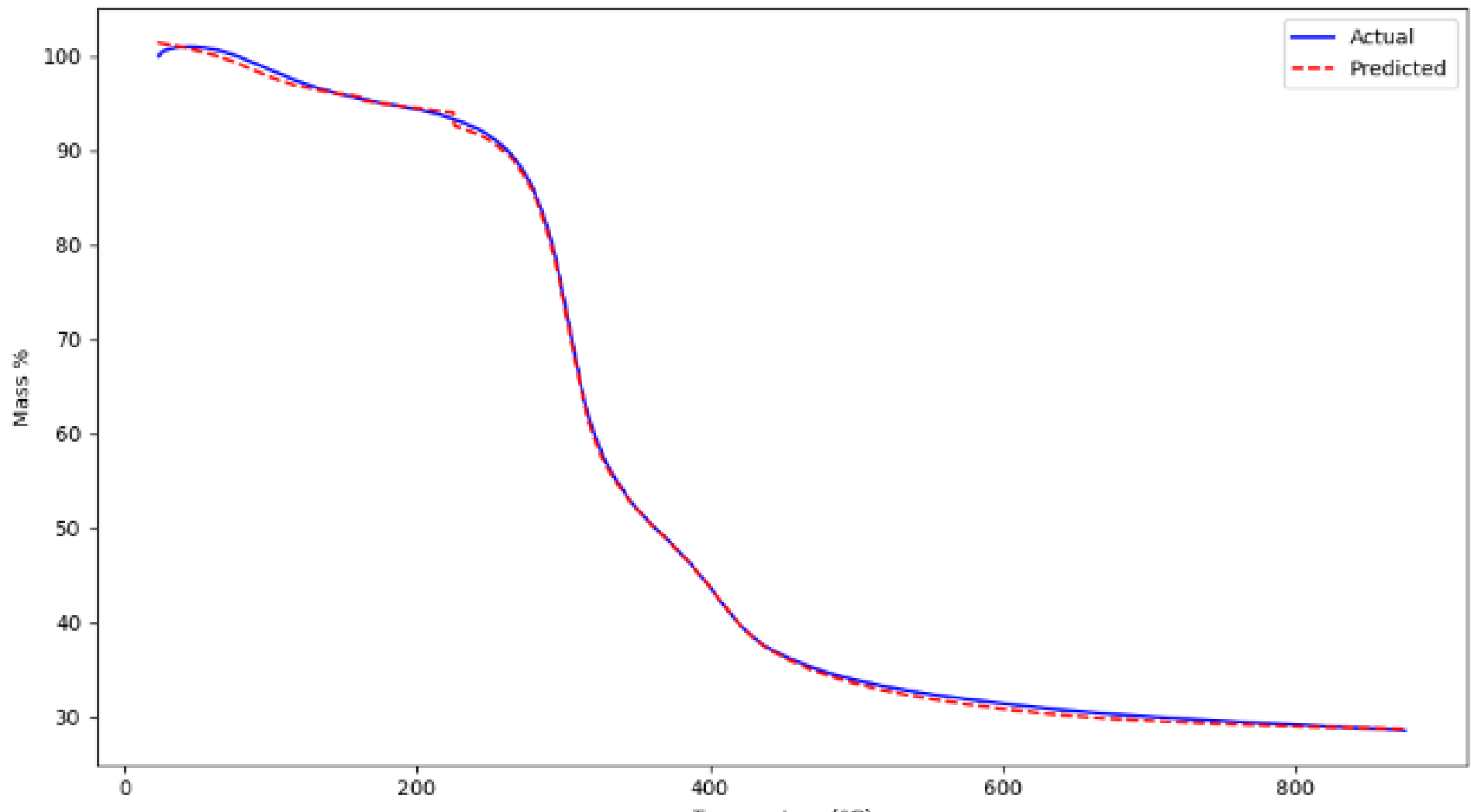

**Figure 21.** Model 2's Prediction of Blend 1's TGA data: (a) 10°C/min; (b) 15°C/min; (c) 25°C/min.

Model 1 (Figure 20) faced challenges in accurately predicting TGA profiles for unseen heating rates and blend ratios. For instance, the model's performance declined at higher heating rates like 15°C/min and 25°C/min (Table 13), which were not included in the training data. This limitation can be attributed to the lack of data diversity, as seen in previous studies [38],[39],[40] where diversified datasets resulted in higher predictive accuracy. Contrarily, Model 2 (Figures 21) addressed this limitation by incorporating lignocellulosic composition features, which provided a deeper understanding of biomass thermal decomposition. As shown in Table 13, this improvement led to significantly lower error metrics (MSE and MAE) and higher $R^2$ values. However, the results indicate that further enhancements, such as increasing dataset diversity, could further boost model performance.

## 4. Conclusions & perspectives

This research emphasizes the importance of shifting from fossil fuel-based "Gray hydrogen" to renewable "green hydrogen" using underutilized biomass resources and machine learning (specifically deep learning) applications. The study explored the thermal, kinetic, and thermodynamic behaviours of Spent Coffee Grounds (SCG) and Date Seeds (DS), and their blends for hydrogen production through pyrolysis process.

The results demonstrate that DS, SCG, and their blends, hold significant promise for bioenergy production. Blend 3 (75% SCG–25% DS) emerged as the most favourable for hydrogen production, with highest volatile matter release. However, it requires highest energy input. Blend 1 (75% DS–25% SCG) demonstrated superior energy efficiency with lowest energy demand. Building on these experimental findings, the study explores the LSTM modelling approach to predict TGA mass loss patterns, significantly outperforming traditional methods, with exceptional accuracy ($R^2 > 0.999$), while considering the lignocellulosic composition of pure biomasses and their blends. These findings demonstrate that combining experimental and predictive approaches can promote the bioenergy production while reducing reliance on time-intensive TGA experiments and pyrolysis reactors.

As perspectives, future work will focus on expanding the diversity of the dataset to other biomasses, and to predict blend behaviour. An attempt to eliminate the need for blend-specific training data is envisaged. Additional efforts will enhance dataset diversity in blend ratios and applied heating rates in order to improve model generalisation.

## Data Availability

Data will be made available on request.